\documentclass[10pt,journal,compsoc]{IEEEtran}

\usepackage[english]{babel}

\usepackage{subfigure}
\usepackage[cite, graphics, square, numbers, comma, latexsym, sort&compress]{}
\usepackage{verbatim}
\usepackage{multirow}
\usepackage{graphicx, epstopdf}
\usepackage{xspace,epsfig,url}
\usepackage{amssymb,amsmath}
\usepackage[noadjust]{cite}
\usepackage{float}
\usepackage{epsf}
\usepackage{psfrag}
\usepackage{verbatim}
\usepackage[all]{xy}
\usepackage{comment}
\usepackage{color,soul}
\usepackage{cases}
\usepackage[noend]{algpseudocode}
\usepackage{array}
\usepackage{multicol}
\usepackage[table]{xcolor}

\title{\LARGE \bf Design Requirements of Generic Hand Exoskeletons and Survey of Hand Exoskeletons for Rehabilitation, Assistive or Haptic Use}

\author{Mine Sarac*, \and Massimiliano Solazzi**, \and Antonio Frisoli**
\thanks{*Department of Mechanical Engineering, Stanford University, Stanford, CA.}
\thanks{{\tt\small msarac@stanford.edu}}%
\thanks{**Perceptual Robotics, Scuola Superiore Sant'Anna in Pisa, Italy.}
\thanks{{\tt\small massimiliano.solazzi@santannapisa.it}}
\thanks{{\tt\small antonio.frisoli@santannapisa.it}}
\thanks{This research was funded by "WEARHAP – WEARable HAPtics for humans and robots" of the European Union Seventh Framework Programme $FP7/2007-2013$ (no $601165$) and "CENTAURO - Robust Mobility and Dexterous Manipulation in Disaster Response by Fullbody Telepresence in a Centaur-like Robot" of the European Union's Horizon 2020 Programme (no $644839$)}
}

\graphicspath{{./Figures/}}

\begin{document}

\maketitle

\begin{abstract}
Most current hand exoskeletons have been designed specifically for rehabilitation, assistive or haptic applications to simplify the design requirements. Clinical studies on post-stroke rehabilitation have shown that adapting assistive or haptic applications into physical therapy sessions significantly improves the motor learning and treatment process. The recent technology can lead to the creation of generic hand exoskeletons that are application-agnostic. In this paper, our motivation is to create guidelines and best practices for generic exoskeletons by reviewing the literature of current devices. First, we describe each application and briefly explain their design requirements, and then list the design selections to achieve these requirements. Then, we detail each selection by investigating the existing exoskeletons based on their design choices, and by highlighting their impact on application types. With the motivation of creating efficient generic exoskeletons in the future, we finally summarize the best practices in the literature.
\end{abstract}

\vspace*{-.5\baselineskip}
\section{Introduction}

Physical rehabilitation is indispensable for the treatment of patients with physical or neurological disabilities~\cite{Kwakkel2004}. Such a therapy mostly focuses on \textit{(i)} increasing the effective range of motion (RoM) of the impaired joints, and \textit{(ii)} repeating activities of daily living (ADLs). Robotic devices can replicate the manual labor of the therapist, while improving patients' motor recovery and functional independence~\cite{Prange2006}.

When it comes to hand disabilities, designing suitable robotic devices is even more challenging due to complex anatomy and high mobility of the hand. To overcome these challenges, designers could simplify hand devices by narrowing mechanical properties for certain tasks and disability levels. Even though stationary state-of-the-art devices address physical rehabilitation effectively~\cite{Dovat2008, Chang2014, Brenosa2011, Ueki2012, Butler2017}, most of them are designed as exoskeletons.

Earlier rehabilitation exoskeletons moved patients' fingers as if grasping an imaginary cup without the actual interaction. Since there is no need for a real interaction, such exoskeletons did not have to free the palm of the hand, or apply high amounts of forces to complete power grasping. Clinical studies revealed the positive impact of realism on motor learning~\cite{Hubbard2009, Bayona2005}, and motivated the designers to fuse rehabilitation and assistive exoskeletons to let users interact with real objects during physical therapy. 


On the other hand, earlier rehabilitation exoskeletons moved patients' fingers repeatedly, while patients sat passively. Since patients have no active role in these exercises, such exoskeletons did not have to be transparent or responsive. After clinical studies showed the positive impact of patients' participation on motor learning~\cite{Krichevets1995, Burke2009}, designers started to create active devices, which can allow patients to perform the desired tasks, and can assist them when needed. Active rehabilitation therapies were then integrated with serious game scenarios to define the desired therapy tasks visually in an entertaining setting. After the integration of serious games became the current trend during physical therapy, rehabilitation and haptic exoskeletons need to get merged to track patients' hand movements and render realistic forces when a virtual interaction occurs.


Even though most of the hand exoskeletons have been designed specifically for a single application, drawing a line between them gets harder every day, and soon will be impossible. 
When it happens, we will need generic exoskeletons that are application-independent. Although technological advances help designers to create better products every day, hand exoskeletons will still suffer from the limitations of hand anatomy and mobility.

However, we believe that we can overcome such anatomical issues and use the technology in the most efficient way only by studying the existing exoskeletons in the literature, and by being inspired from the best practices. Rehabilitation exoskeletons have been surveyed before focusing on various categories~\cite{Troncossi2016}, or specific issues, such as actuator technologies~\cite{Heo2012}, or control strategies~\cite{Meng2017}. These surveys effectively reflect how much rehabilitation exoskeletons evolved over time, but they do not focus on future generic devices or implementing current devices for other applications.

%

\begin{figure*}[t!]
\centering
  \subfigure[Rehabilitation exoskeletons~\cite{Tong2010} \label{fig:rehab}]%
	{\includegraphics[width=0.26\textwidth]{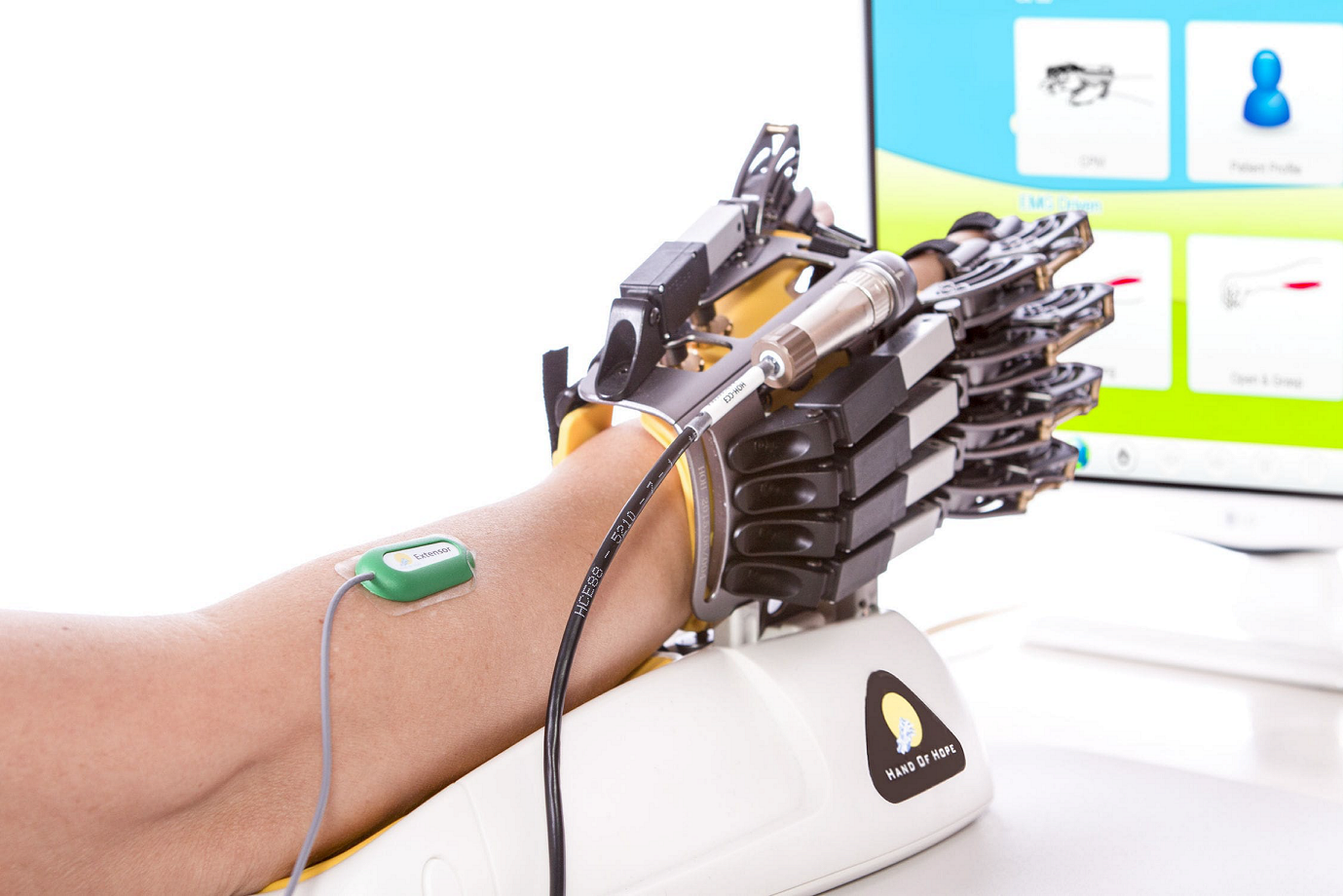}} \hspace{0.03\textwidth}
  \subfigure[Assistive exoskeletons~\cite{Gasser2017} \label{fig:assist}]%
	{\includegraphics[width=0.21\textwidth]{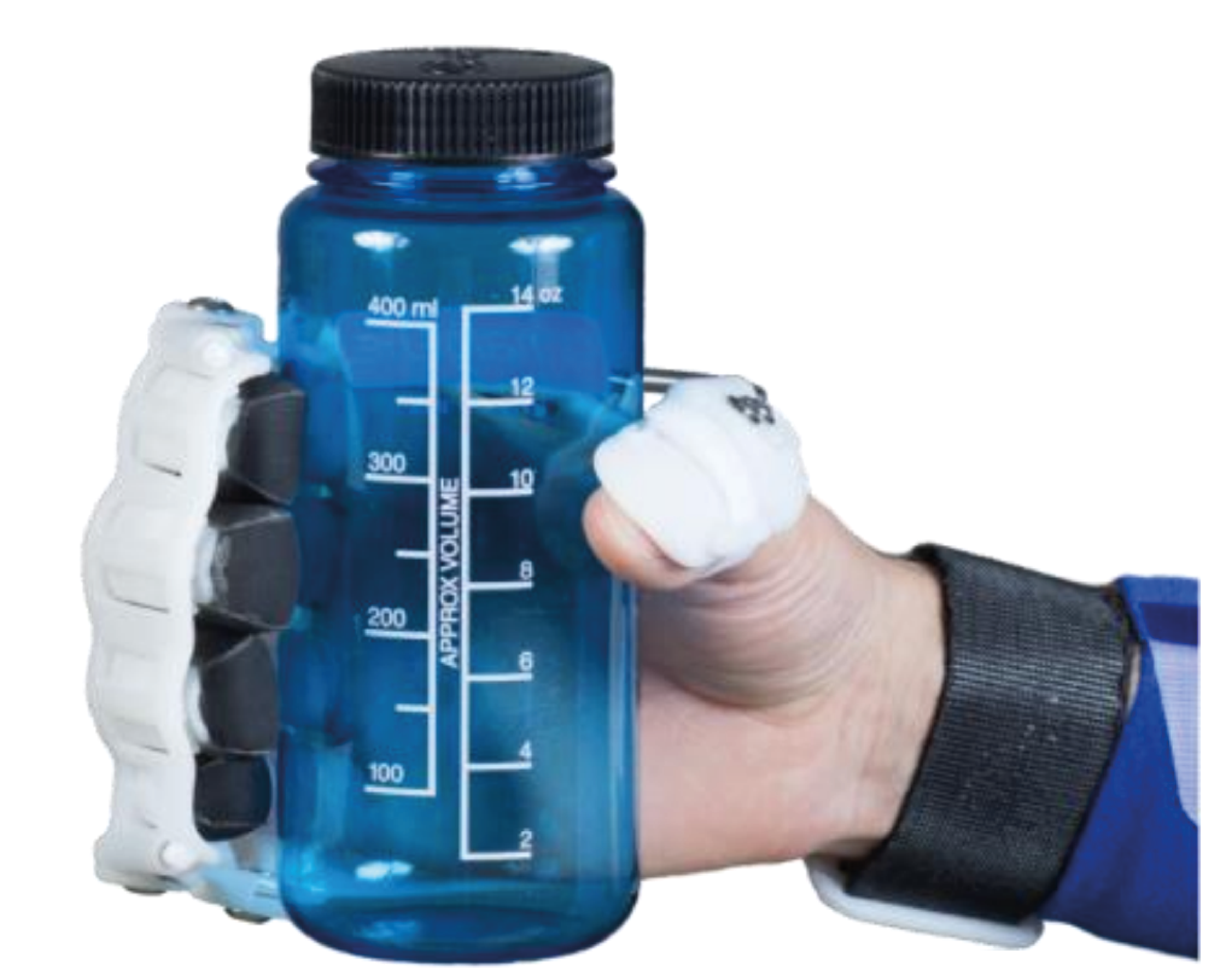}} \hspace{0.03\textwidth}
	\subfigure[Haptic exoskeletons~\cite{Choi2016} \label{fig:hapt}]
	{\includegraphics[width=0.26\textwidth]{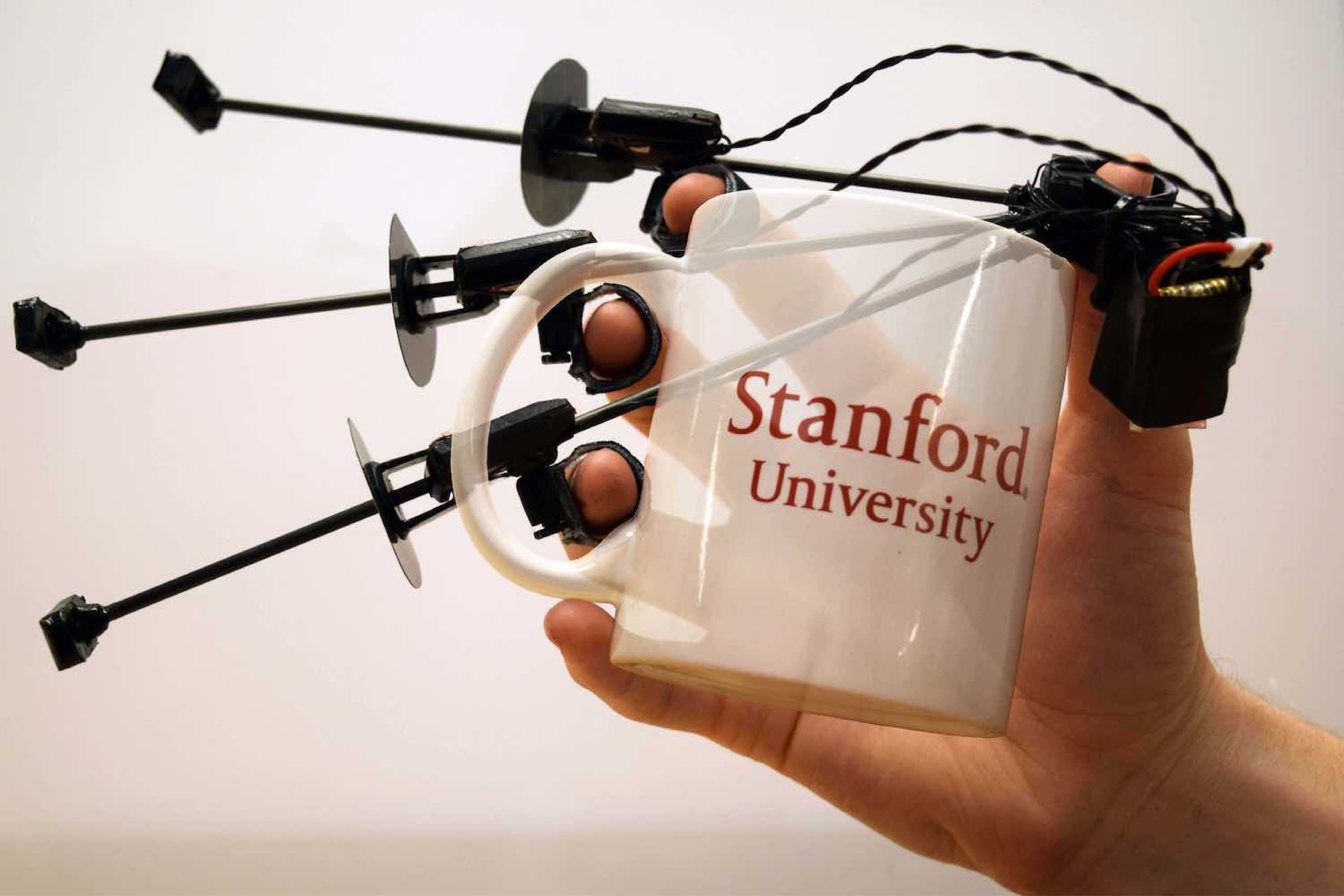}} \hspace{0.03\textwidth}
\vspace*{-.75\baselineskip}
  \caption{Generic hand exoskeletons should be designed to be operational for different applications: (a) rehabilitation, (b) assistive and (c) haptic.}
	\label{fig:applications}
\vspace*{-1\baselineskip}
\end{figure*}

In this paper, we aim  at presenting a systematic guideline, and listing the best practices of generic hand exoskeletons for future designers. First, we will define each target application, and list the general and application-specific properties of hand exoskeletons. For each application-specific property, we will highlight with which design selections it can be achieved. Once we have the design selections, we will start investigating the hand exoskeletons in the literature to reveal all possible choices and discuss them from the perspective of target applications~\footnote{This study focuses on hand exoskeletons developed between $2002$ and $2018$}. Finally, we will summarize the best design practices as a guideline for future designers.

\vspace*{-0.5\baselineskip}
\section{Design Requirements for Exoskeletons} \label{sec:design}

A hand exoskeleton is a wearable device that provides realistic kinesthetic feedback to user's fingers through active force transmission over a series of mechanical components. There are different properties a hand exoskeleton has to satisfy because of either human interaction, or the target application. In this section, we will first list these generic requirements and then study specific requirements by highlighting which decisions a designer can take to satisfy them.

\vspace*{-0.3\baselineskip}
\subsection{General properties for all applications} \label{sec:prop1}

\begin{figure}[h!]
  \centering
  \vspace*{-.5\baselineskip}
  \resizebox{1.3in}{!}{\includegraphics{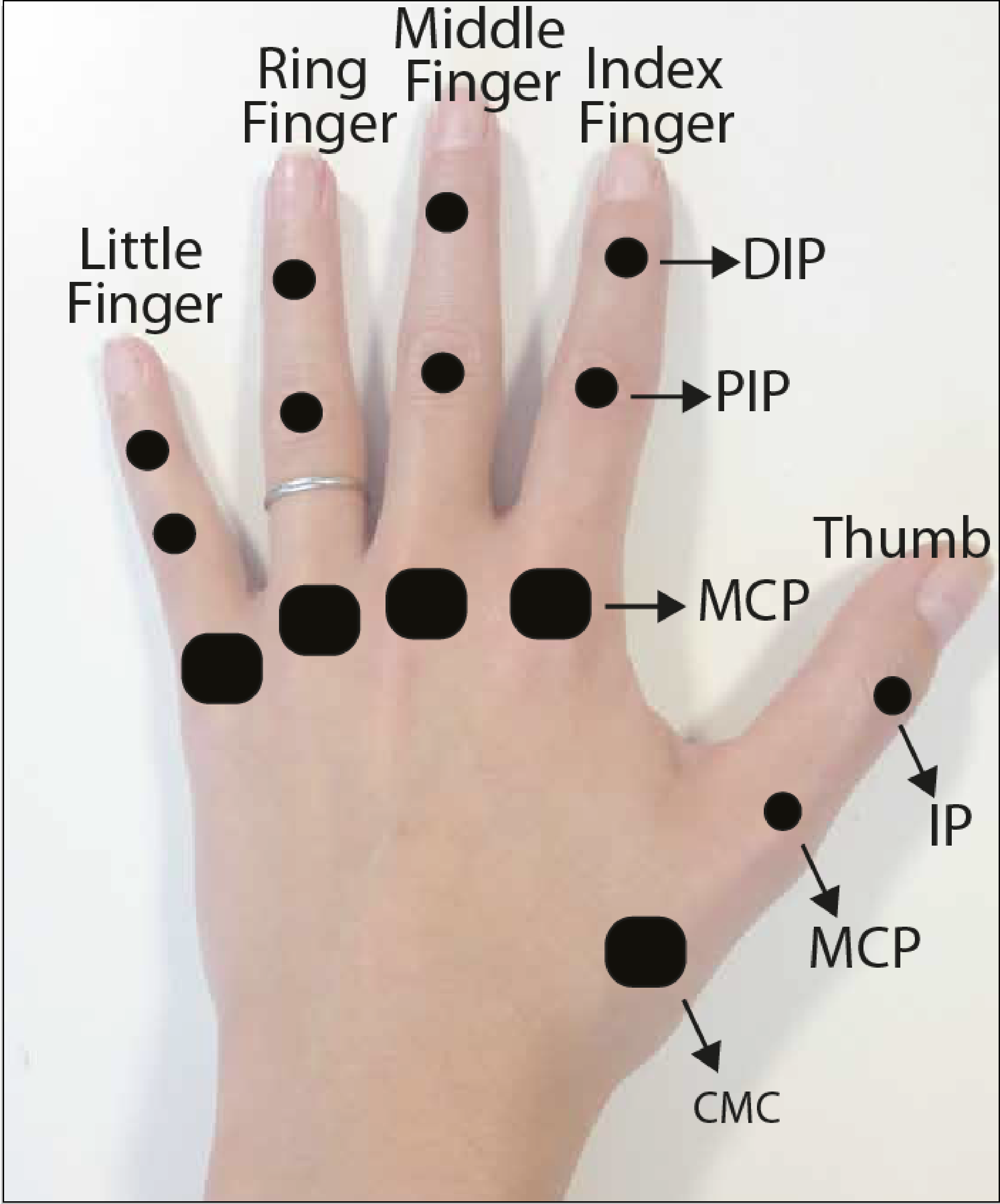}}
  \vspace*{-.5\baselineskip}
  \caption{Kinematic model of a hand: each finger has $3$ joints with $4~DoF$: index, middle, ring and little fingers have MCP, PIP and DIP joints, thumb has CMC, MCP and IP joints.}
  \label{fig:hand}
\end{figure}

\textbf{- Hand anatomy:} Human hand has $5$ fingers, $15$ joints and $20$ degrees of freedom (DoF) mobility~\cite{Buchholz1992}, and a hand exoskeleton must comply with the anatomy of the hand (see Figure~\ref{fig:hand}). Index, middle, ring and little fingers have $3$ joints with $4~DoF$: metacarpophalangeal (MCP) with $1~DoF$ flexion/extension and $1~DoF$ abduction/adduction, proximal interphalangeal (PIP) and distal interphalangeal (DIP) with $1~DoF$ flexion/extension each. Even if DIP and PIP are physically independent, they are anatomically coupled to each other, so they move together. Most of the ADLs require only flexion/extension, while abduction/adduction adjusts the hand posture. Similarly, thumb has $3$ joints with $4~DoF$: carpometacarpal (CMC) with $1~DoF$ flexion/extension and $1~DoF$ abduction/adduction, MCP and Interphalangeal (IP) with $1~DoF$ flexion/extension each.

\noindent \textbf{- Safety:} A hand exoskeleton must ensure user's safety at all times. The mechanical and control systems of the exoskeleton must respect the natural movements of finger joints~\cite{Buchholz1992} and hand size~\cite{Buryanov2010}. Furthermore, mechanical limits must ensure not to exert forces to finger joints once they reach the joint limits.

\noindent \textbf{- Comfort:} A hand exoskeleton must be comfortable for the user, as the user must be wearing the device during operation. Kinematics and ergonomic design of the device must ensure not to cause any pain or fatigue.


\noindent \textbf{- Effective force transmission:} A hand exoskeleton must transmit actuator forces to user's finger naturally. While controlling multiple finger joints, torques around finger joints must have a balanced ratio not to cause pain at any orientation. Finally, forces between the exoskeleton and finger phalanges must be perpendicular, since tangential forces might cause finger connections to slip from the finger.

\noindent \textbf{- Affordability:} A hand exoskeleton must be affordable for therapy clinics, so patients could afford utilizing these devices for their rehabilitation process. They should require low maintenance and easy to be used by non-technical staff. Furthermore, these devices should be functional for a wide range of patients in terms of hand sizes, or disability levels.

\vspace*{-0.3\baselineskip}
\subsection{Specific properties for different applications} \label{sec:prop2}

Designers can choose a specific target application to design simpler mechanisms, since each application has specific desired tasks, user profiles and external factors.

\vspace*{0.5\baselineskip}
\textbf{Rehabilitation exoskeletons} are designed to treat disabilities of patients in a clinical setting (see Figure~\ref{fig:rehab}). These devices focus on repetitive therapy tasks, which mostly mimic the most common ADLs by opening/closing the fingers. Rehabilitation exoskeletons must be easily wearable not to cause discomfort or pain for the patients during preparation. These devices are preferred to allow patients to interact with real objects, to apply high output forces, and to monitor finger movements for performance evaluation. Instant adjustability for different tasks also is favorable, even though patients with severe disabilities would not take advantage of the task variety due to the loss of isolated individual finger movement after injury~\cite{Welmer2008}. Their portability is not mandatory especially for clinical devices, but still preferable.

\textbf{Assistive exoskeletons} are designed to assist patients with hand disabilities in performing ADLs in their daily life, such as grasping a cup while drinking coffee, or holding a key while opening the door (see Figure~\ref{fig:assist}). Instant adjustability for different tasks, easy wearability and portability are highly important for assistive exoskeletons. They must allow patients to interact with real objects, and to apply high output forces. Finger tracking is neither mandatory, nor favorable.

\textbf{Haptic exoskeletons} are designed for healthy subjects to interact with a virtual environment (see Figure~\ref{fig:hapt}). Instant adjustability for different tasks, portability and efficient finger tracking are highly important for haptic exoskeletons. Since the target user profile is assumed to be healthy, the wearability or the amount of output forces are not mandatory like other applications but favorable.

Based on the definitions of target applications, we would like to detail these specific properties. Since each property can be achieved through design choices, the appropriate design selections will be listed:

\noindent \textbf{- Independent finger control:} Grasping different objects require fingers to move in different synergies. To assist users in grasping different objects, assistive and haptic exoskeletons must control fingers independently. Rehabilitation exoskeletons can offer repetitive physical therapy by controlling fingers together or by opening/closing the hand in a unique way. Even though independent finger control is not mandatory, it is favorable.

\noindent Design selections: \emph{hand mobility}

\noindent \textbf{- Grasping objects with generic shapes:} Gasping different objects also require finger joints to move in different synergies, e.g. picking, grasping or scooping~\cite{Thea1997}. To assist users in grasping different objects, allowing finger joints to move independently is mandatory for assistive and haptic exoskeletons, but favorable for rehabilitation exoskeletons.

\noindent Design selections: \emph{finger mobility}

\noindent \textbf{- Easy wearability:} Wearing the hand exoskeleton can be much more painful and harder for patients with hand disabilities compared to healthy users. Rehabilitation and assistive exoskeletons must ensure the exoskeleton to be worn easily and without a strict initial pose. Yet, wearability is not a as crucial for haptic exoskeletons.

\noindent Design selections: \emph{number of interaction points}, \emph{kinematics selection} and \emph{adjustment for hand sizes}

\noindent \textbf{- Interaction with Real Objects:} Assistive exoskeletons must allow patients with disabilities to interact with real objects. Interaction with real objects is also favorable for rehabilitation exoskeletons to increase the realism perception of therapy tasks, but not mandatory. On the other hand, it is neither mandatory nor favorable for haptic exoskeletons, since they are designed for virtual interactions.

\noindent Design selections: \emph{mechanism placement}

\noindent \textbf{- High output forces:} Patients with disabilities suffer from high stiffness along their joints, so rotating their finger joints requires higher output forces than rotating finger joints of healthy users. Rehabilitation and assistive exoskeletons must apply high output forces to move finger joints with high stiffness, while haptic exoskeletons can apply relatively lower forces to render virtual interaction forces.

\noindent Design selections: \emph{actuator selection} and \emph{direction of movement}

\noindent \textbf{- Portability $\&$ lightness:} Assistive and haptic exoskeletons must be portable and lightweight to allow users explore real or virtual environment. On the other hand, rehabilitation exoskeletons can be used in a clinic setting, so their portability is not as crucial as others but favorable.

\noindent Design selections: \emph{transmission system}

\noindent \textbf{- High transparency:} Assistive and haptic exoskeletons must be backdriveable, so that users can move freely in real/virtual environments. For rehabilitation, backdriveability is highly favorable especially to ensure user's safety and to let patients participate the therapy tasks, but is not mandatory.

\noindent Design selections: \emph{actuator selection}, \emph{transmission system} or \emph{control system}

\noindent \textbf{- Finger Pose:} Haptic exoskeletons must track user's movements and reflect them into virtual environment. Rehabilitation exoskeletons should track user's movements to monitor their performance improvement during therapy, or to implement virtual game scenarios. On the other hand, finger pose is neither mandatory not favorable for assistive exoskeletons.

\noindent Design selections: \emph{finger pose tracking}.

\begin{figure}[h!]
  \centering
  \vspace*{-1\baselineskip}
  \resizebox{3.2in}{!}{\includegraphics{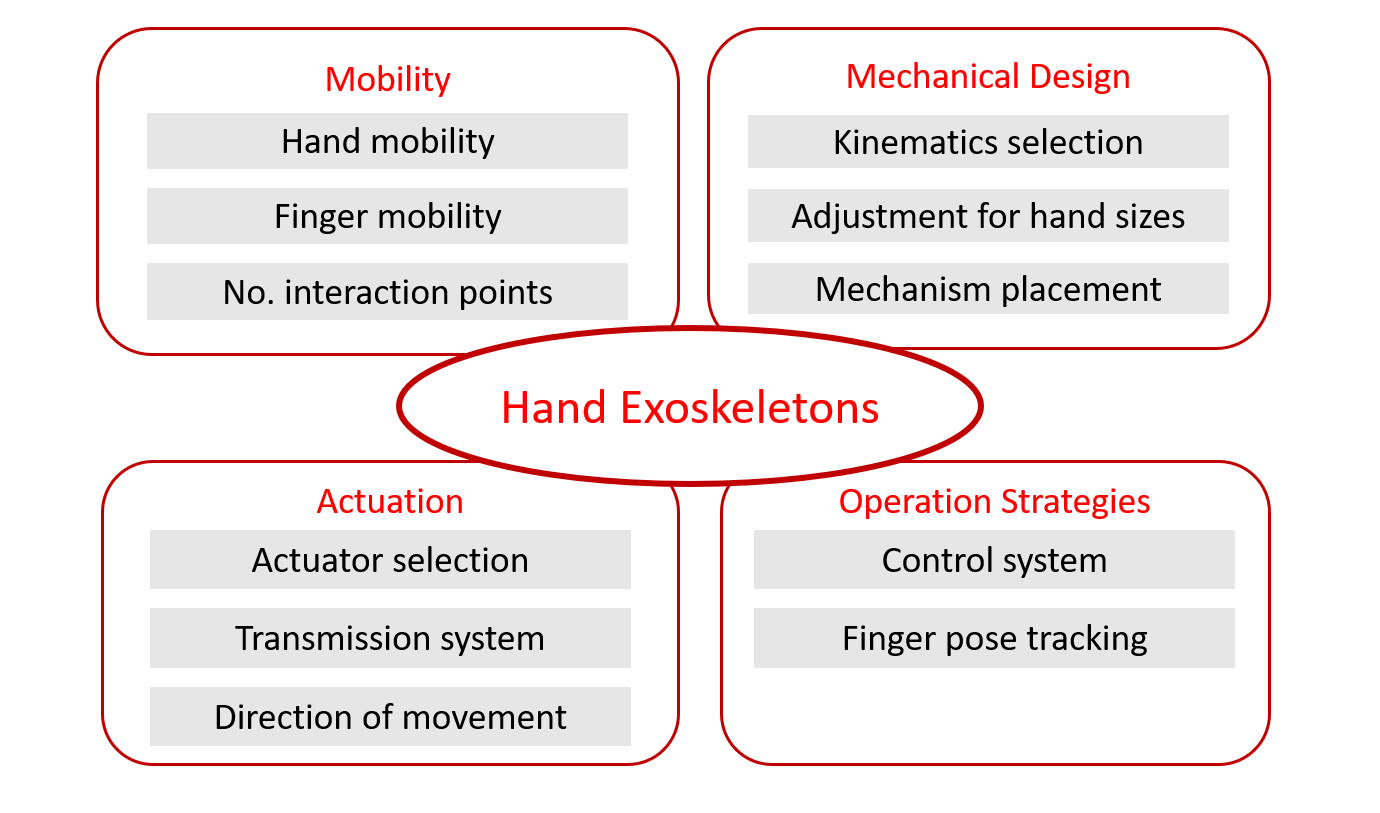}}
  \vspace*{-.75\baselineskip}
  \caption{Hand exoskeletons should be categorized by design selections that can be categorized under design aspects: mobility, mechanical design, actuation and operational strategies.}
  \label{fig:structure}
  \vspace*{-1\baselineskip}
\end{figure}

These design selections are chosen to satisfy the corresponding properties, and can affect each other directly or indirectly. Therefore, these selections should be investigated not individually, but based on certain aspects, which are formed by combining multiple selections. Figure~\ref{fig:structure} represents these design aspects, such as mobility, mechanical design, actuation and operation strategies, and the design selections under each aspect.

\vspace*{-0.3\baselineskip}
\subsection{Assumptions for the exoskeletons} \label{sec:assumptions}

In the next section, we will analyze the state-of-the-art hand exoskeletons based on design selections with the motivation of creating generic exoskeletons achieving all the aforementioned properties. Some of these properties are based on quantitative data (e.g. hand mobility, finger mobility, etc.), or technical robotic background (e.g. finger pose, high transparency). Even though others (e.g. overall cost, portability, lightness, etc.) should be based on quantitative data, most of the publications suffer from the lack of details in this matter. This is why we will generalize exoskeletons based on qualitative inferences.

We will label exoskeletons as:

\noindent \textbf{- \emph{portable}} if all actuation and sensing units are mounted on the exoskeleton, while its controller or power units can be placed remotely on a tabletop, and can be connected to the exoskeleton through a single cable;  \\
\textbf{- \emph{light}} if either a single miniaturized actuator is mechanically attached for each finger component, or multiple actuators or differential transmission system are equipped to the finger components remotely; \\
\textbf{- \emph{low-cost}} if each finger component is controlled by a single actuator, since actuators are usually the most expensive parts for an exoskeleton; \\
\textbf{- \emph{easily wearable}} if the exoskeleton is composed of compliant links, or rigid links with passive joints, and the links are connected to user's fingers with adjustable straps; or \\
\textbf{- \emph{natural}} if the finger components do not force the user's fingers to move in a strict predefined manner, and the exoskeleton is actively or passively backdriveable.

\vspace*{-0.5\baselineskip}
\section{Hand Exoskeletons in the Literature} \label{sec:literature}

State-of-the-art exoskeletons will be investigated based on their design choices for each selection, as summarized in Figure~\ref{fig:structure}. We will then discuss whether each possible design choice is suitable for a generic hand exoskeleton. For further reference to the user, the full list of the exoskeletons studied in this paper has been listed in Table~\ref{tab:5fin} and Table~\ref{tab:4321fin}.

\vspace*{-0.3\baselineskip}
\subsection{Mobility}

\begin{figure}[b!]
  \centering
  \vspace*{-1.75\baselineskip}
  \resizebox{3.2in}{!}{\includegraphics{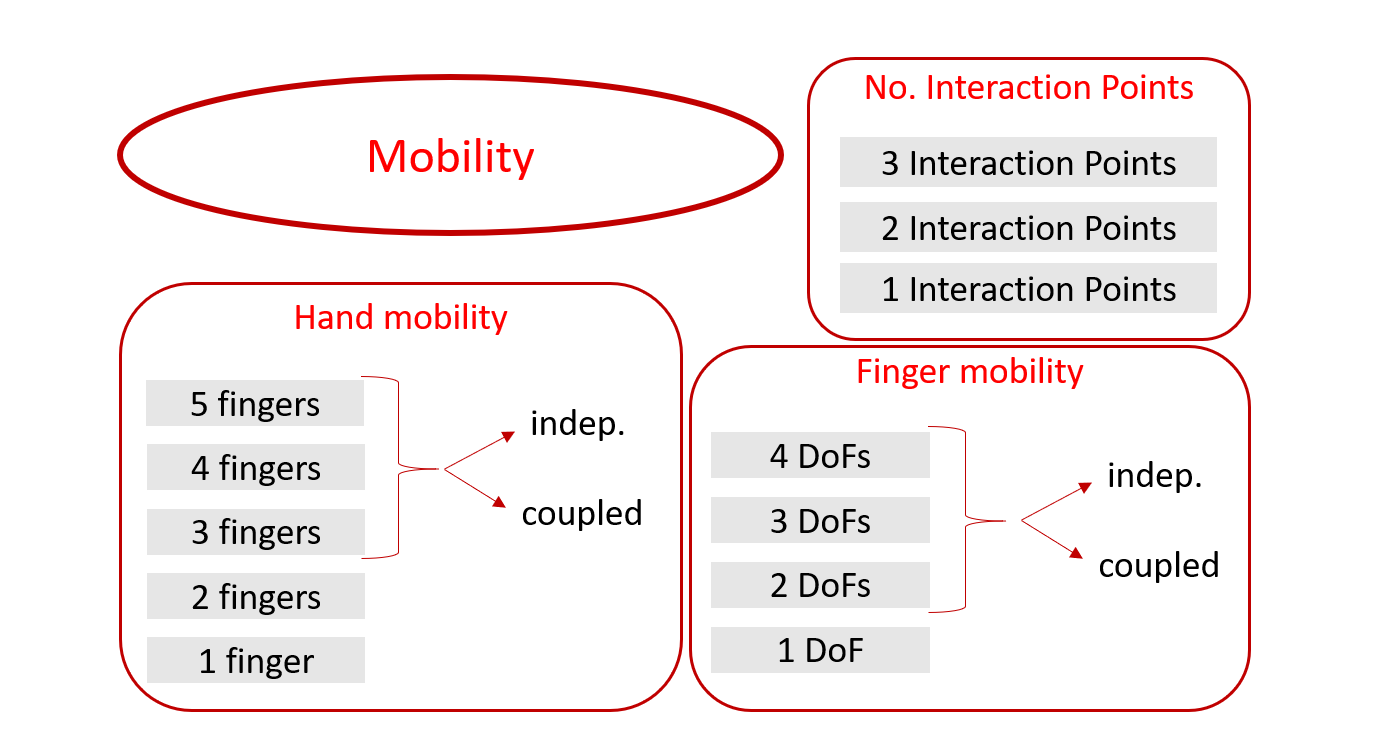}}
  \vspace*{-.75\baselineskip}
  \caption{Possible design choices for mobility based on hand mobility, finger mobility and number of interaction points.}
  \label{fig:mobility}
\end{figure}

Mobility assisted by an exoskeleton can be handled in terms of hand mobility, finger mobility and the number of interaction points between the mechanism and user's finger (see Figure~\ref{fig:mobility}). Both hand and finger mobilities can be categorized further based on the number of assisted and independently controlled mobility.

\vspace*{-0.3\baselineskip}
\subsubsection{Hand mobility} \label{sec:q1q2}

A human hand has $5$ fingers, and an exoskeleton can be designed to assist and control various numbers of fingers. Finger exoskeletons~\cite{Aubin2013, Maeder-York2014, Hocaoglu2009, Sun2009, Lee2015, Jones2010, Polotto2012, Wang2009, Agarwal2013, Tang2011, DiCicco2004, Wege2005, Yamaura2009} are designed mostly for the index finger, and are mostly stated as an initial study for a multi-finger exoskeleton. $2$-finger exoskeletons control thumb and index finger independently, and support only specific hand movements for rehabilitation or haptics, such as finger tapping or pick-and-place tasks~\cite{Cempini2013, Cempini2015, Taheri2014, Gosselin2005, Fontana2009, Li2011, Fiorilla2008, Cortese2015, Hasegawa2011}.

Even though exoskeletons with $1$ or $2$ fingers are simpler to implement, most of ADLs require at least $3$ fingers to be assisted. One approach to design multi-finger exoskeletons is to control each finger component individually. $5$-finger exoskeletons control each finger independently, and can be used for all applications with minimum constraints~\cite{Burton2011, Fu2007, Tong2010, Chiri2009, Rahman2012, Cui2015, Kim2017, Connelly2010, Delph2013, Decker2017, Jo2017, Yap2015, Polygerinos2015, BenTzvi2015, Fang2009, Lu2016, Jo2014, Iqbal2011, Sarac2016}. Since the middle, ring and little fingers of a healthy person are highly coupled, $4$-finger exoskeletons, which control thumb, index, middle and ring fingers~\cite{Popov2017, Bouzit2002}, or $3$-finger exoskeletons, which control thumb, index and middle fingers~\cite{Kobayashi2012, Ryu2008, Sarakoglou2016, Wei2017}, can be used all applications. Even though these devices can assist users during all ADLs, the perception of realism would drop as the number of assisted fingers decrease. On the other hand, $4$-finger exoskeletons, which control index, middle, ring and little fingers, cannot be effective for grasping or picking tasks during assistive or haptic applications due to the lack of resistive forces acting on the objects through the thumb~\cite{Allotta2015}.

Increasing the number of assisted fingers improves the overall mobility while complicating the design. The second approach to design multi-finger exoskeletons is to couple finger movements through mechanical~\cite{Hasegawa2008, Ferguson2018, Burton2011, Mulas2005, Lee2013, Brokaw2011, Troncossi2012, Kobayashi2013, Wu2010, Wei2013, Gasser2015, Lince2017} or differential~\cite{Li2017, Arata2013, Weiss2013, In2010, Geo2016, Gasser2017} systems.

Even though we cannot claim that moving finger components together prevents the exoskeleton to be used for certain applications, it limits certain tasks. For instance, a $5$-finger exoskeleton with coupled index, middle, ring and little fingers can assist users grasping objects only in certain shapes (e.g. a water bottle) during assistive or haptic applications, but not a key. This is why a generic hand exoskeleton should control $4$ or $5$ fingers independently.


\begin{figure*}[t!]
\centering
  \subfigure[$3~Points$ ($3~DoF$)~\cite{Wege2007} \label{fig:3cp}]%
	{\includegraphics[width=0.19\textwidth]{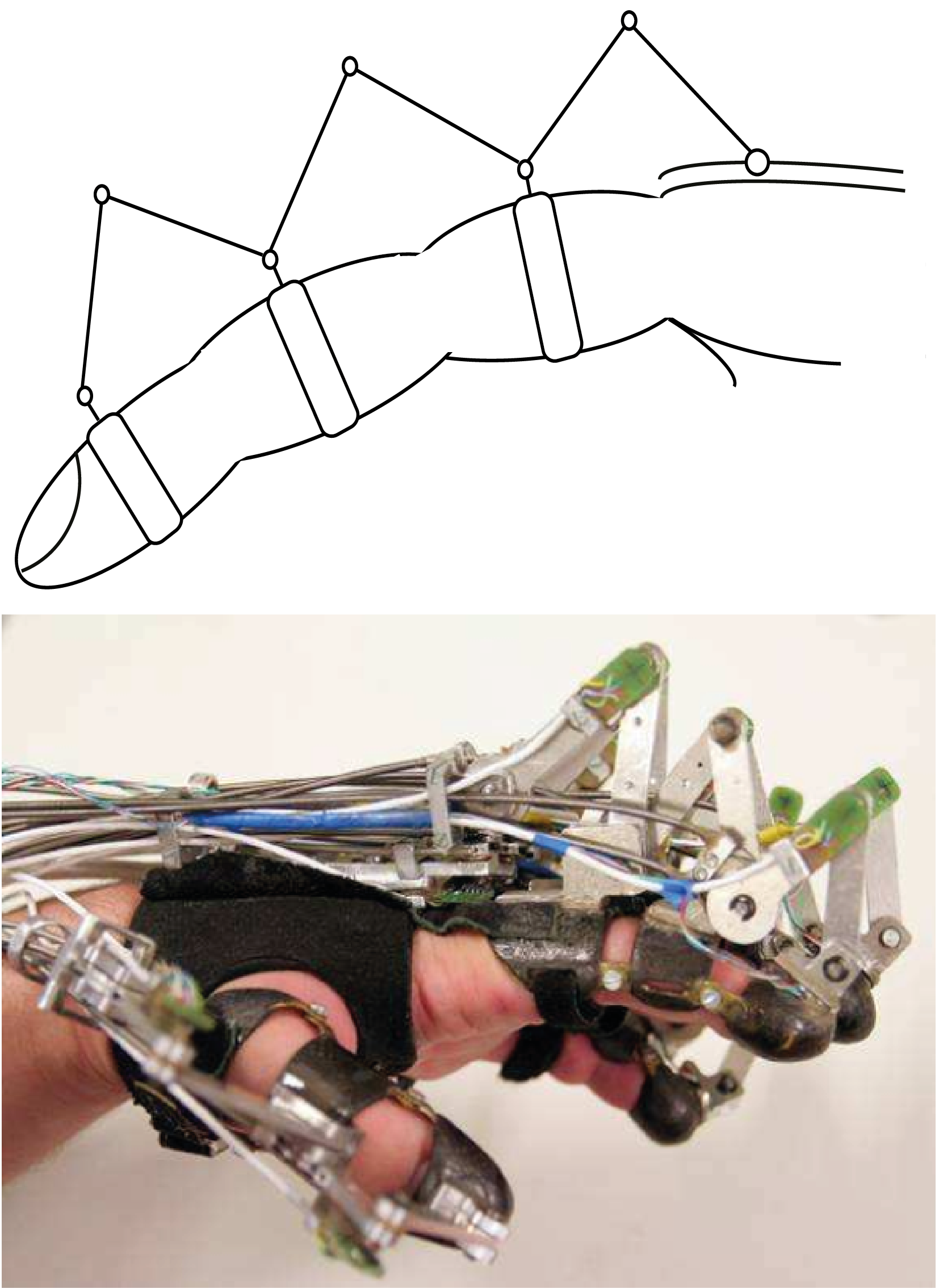}} \hspace{0.01\textwidth}
  \subfigure[$2~Points ($2~DoF$)$~\cite{Tong2013} \label{fig:2cp}]%
	{\includegraphics[width=0.19\textwidth]{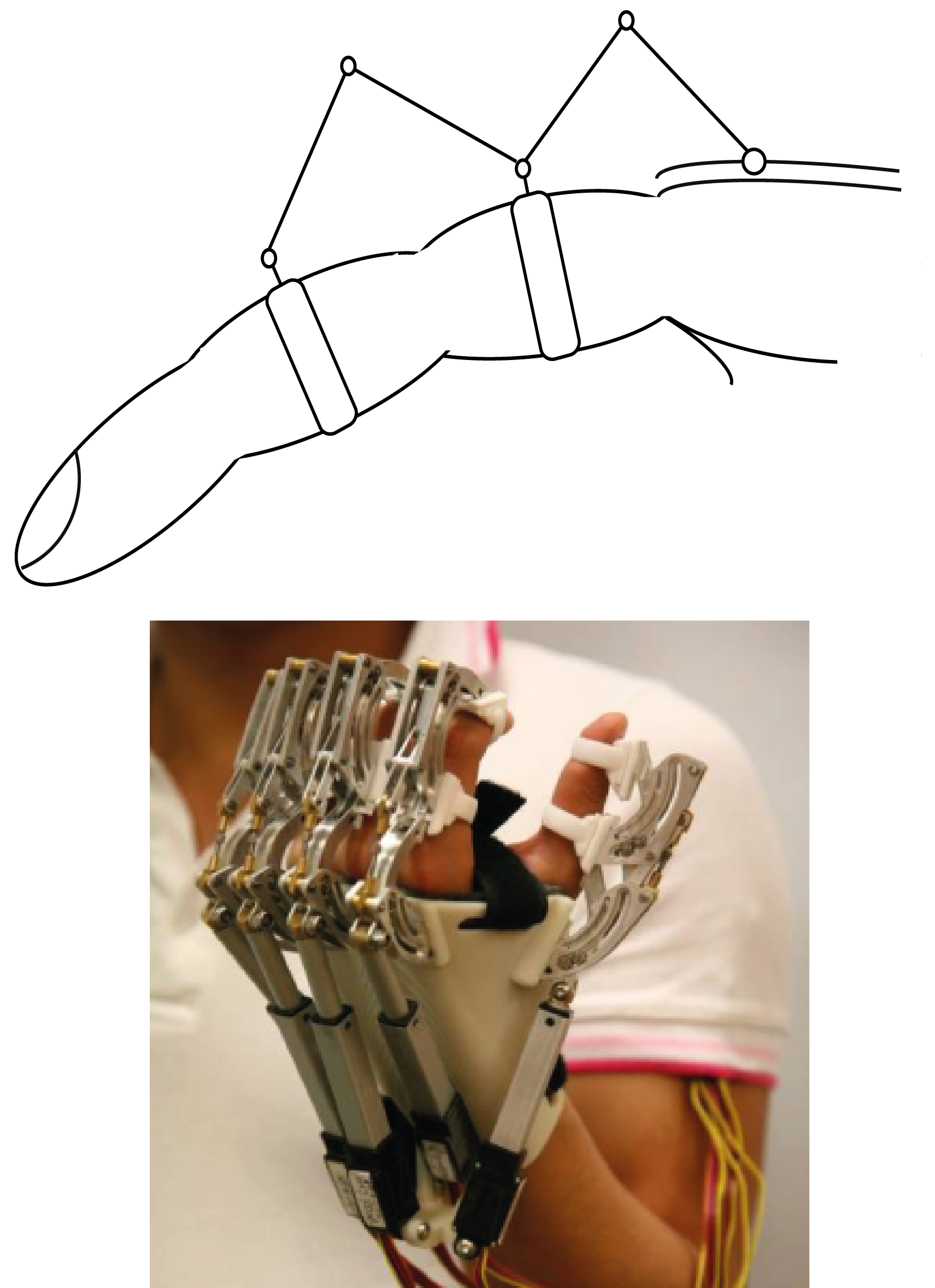}}\hspace{0.01\textwidth}
	\subfigure[$1~Point$ ($1~DoF$)~\cite{Fiorilla2008} \label{fig:1cp_1}]%
	{\includegraphics[width=0.19\textwidth]{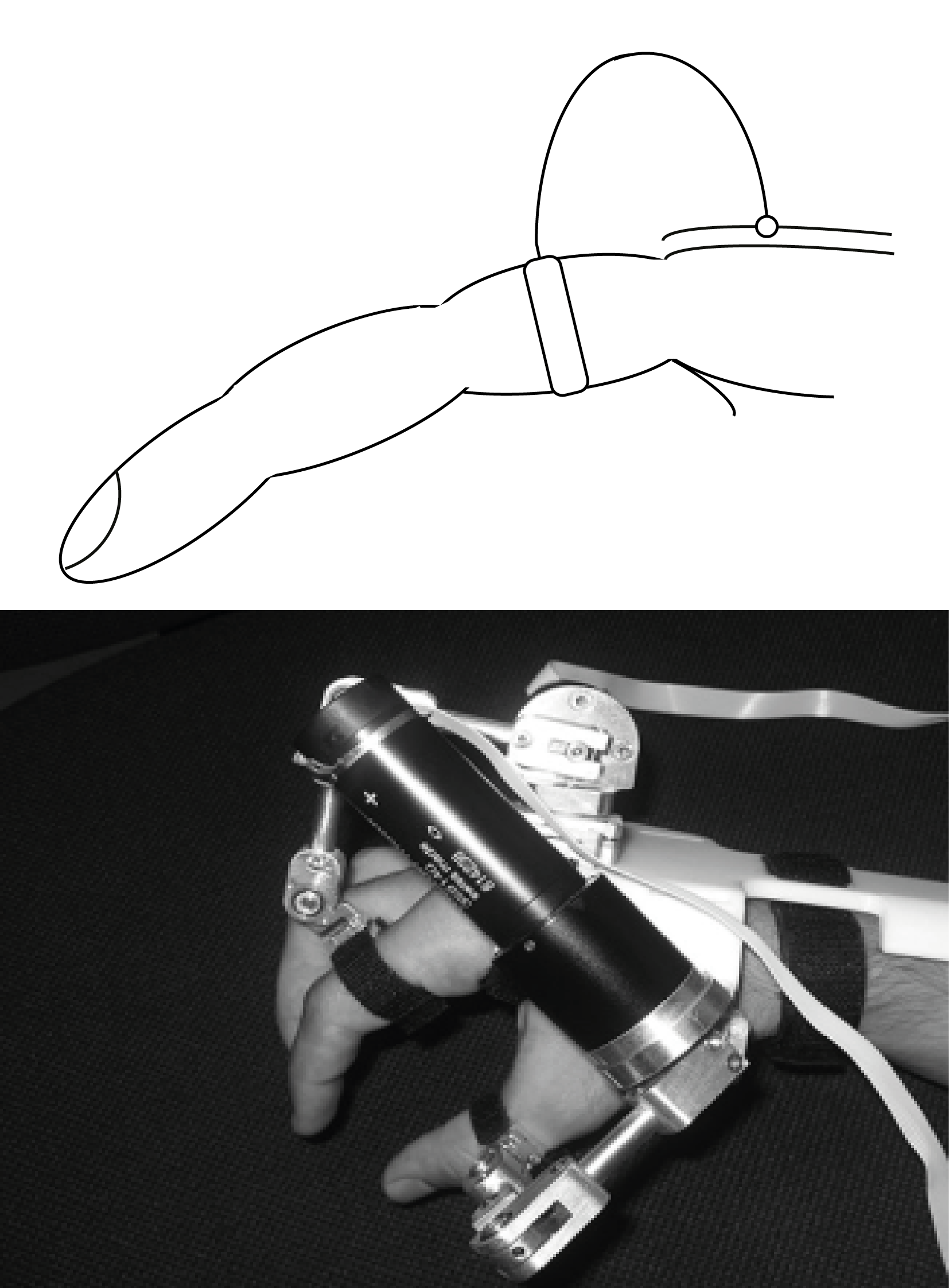}}\hspace{0.01\textwidth}
  \subfigure[$1~Point$ ($3~DoF$)~\cite{Chang2014} \label{fig:1cp_3}]%
	{\includegraphics[width=0.25\textwidth]{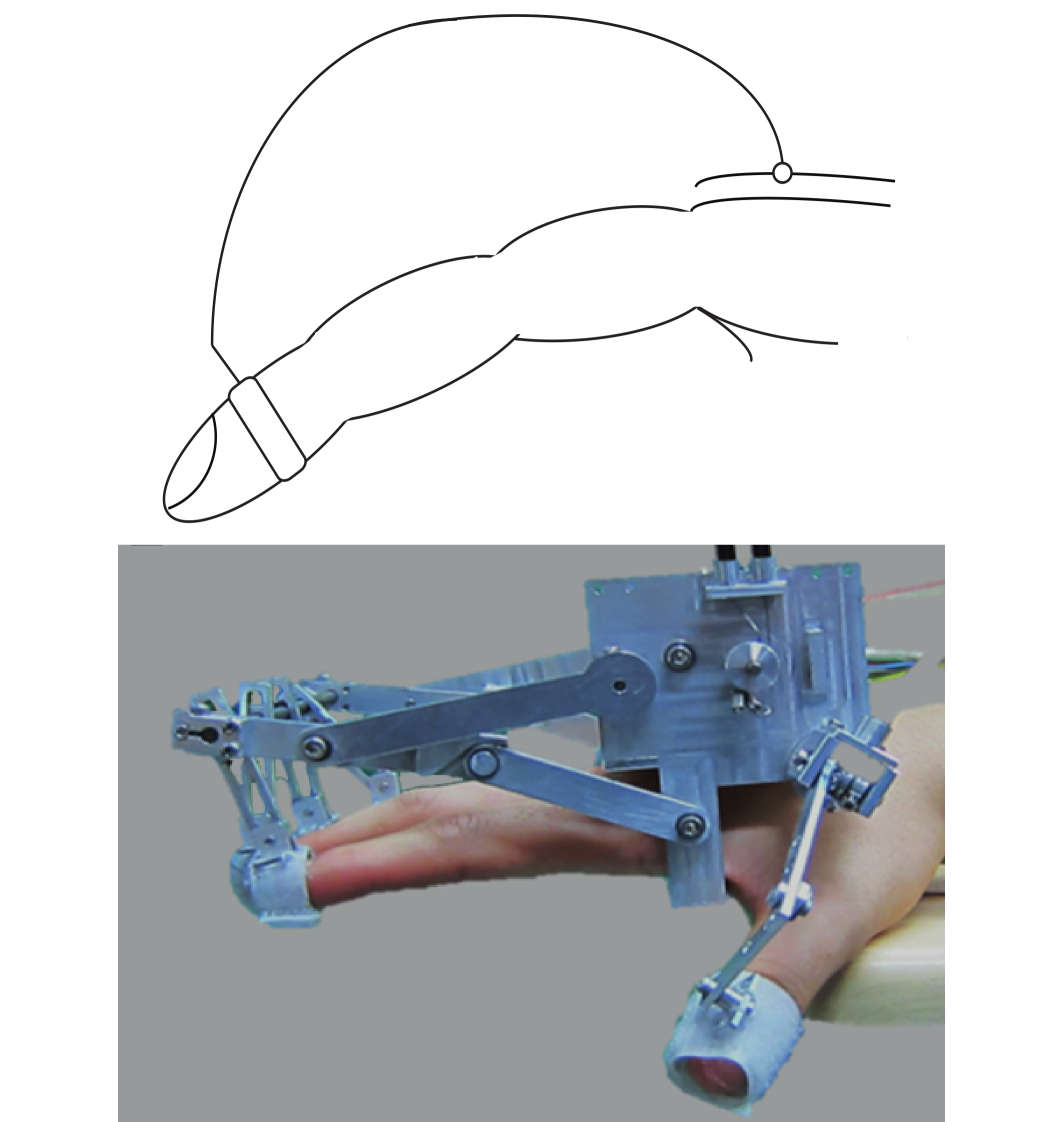}}\hspace{0.01\textwidth}
  \vspace*{-.5\baselineskip}
	\caption{Hand exoskeletons can be designed with different numbers of interaction points between the device and user's fingers. Multiple interaction points improve grasping stability, user's safety and perception of touch but are harder to be worn.}
	\label{fig:cp}
  \vspace*{-.5\baselineskip}
\end{figure*}

\subsubsection{Finger mobility} \label{sec:q3q4}

A human finger has $4~DoF$ mobility, and an exoskeleton can be designed to assist and control various numbers of finger joints for each finger. $1~DoF$ mechanisms~\cite{Brokaw2011, Fiorilla2008} only flex/extend MCP joint for repetitive rehabilitation exercises and enhanced motor learning. Even though finger components with $1~DoF$ mobility are simpler to implement and easier to be worn, most of ADLs require at least $2$ DoF to be assisted for each finger. One approach to design multi-DoF mechanisms is to control each finger joint individually with $2~DoF$~\cite{Aubin2013, Taheri2014, Wu2010}, $3~DoF$~\cite{Jones2010, Hasegawa2008} or $4~DoF$~\cite{Wang2009, Li2011, Polotto2012} mobility.




Increasing the number of assisted joints improves the overall mobility while complicating the design. The second approach to design multi-DoF mechanisms is to couple finger joints through mechanical or differential systems. Towards simplifying the finger components, the first step can be leaving the abduction/adduction of MCP joint passive~\cite{Agarwal2013}, or neglected completely, since most of the ADLs focus on finger opening/closing. Even then, controlling $3~DoF$ flexion/extension independently can be challenging. As the second simplification step, DIP and PIP joints can be coupled with a mechanically adjustable ratio, while MCP joint is controlled independently~\cite{Fontana2009, Ferguson2018, Yamaura2009, Burton2011, Wei2017, Cortese2015, Kim2017, Agarwal2015, Kobayashi2012, Kobayashi2013}. Since DIP and PIP joints are anatomically coupled, this simplification does not affect the perception significantly, but coupling them with a constant ratio might limit certain finger synergies.

Finally, a mechanism can be designed with a single actuator to control finger opening/closing through $4~DoF$~\cite{Bouzit2002, Iqbal2011, Sun2009, Sarakoglou2016, Fu2007, Wege2005, Chiri2009, Weiss2013, Cempini2013, Lee2013}, $3~DoF$~\cite{In2010, Connelly2010, Delph2013, Li2017, Popov2017, Mulas2005, Kobayashi2012, Jo2014, Arata2013, Yap2015, Polygerinos2015, Ryu2008, Gosselin2005, BenTzvi2015, Fang2009, Lince2017, Troncossi2012, Hocaoglu2009, DiCicco2004, Rahman2012, Allotta2015, Cui2015, Lee2015, Geo2016, Maeder-York2014} or $2~DoF$~\cite{Decker2017, Jo2017, Tong2010, Tang2011, Lu2016, Wei2013, Gasser2015, Sarac2016} mobility. Such coupling can set by a constant ratio through mechanical linkages or differential systems, or by adjusting the transmitted forces automatically based on contact forces~\cite{Gosselin2003}.

Even though we cannot claim that moving finger joints together prevents the exoskeleton to be used for certain applications, it limits certain tasks. For instance, a $3~DoF$ mechanism with constant ratio can assist users grasping objects in certain shapes (e.g. a water bottle) during assistive or haptic
applications, but not a phone without having mechanical adjustments. This is why a generic hand exoskeleton should flex/extend $2$ or $3$ finger joints independently, or coupled based on contact forces. Compared to fully controlled mechanisms, underactuated systems based on contact forces are mechanically simpler and cheaper, but require more complicated operational strategies.



\subsubsection{Number of interactions} \label{sec:q5}

A human finger has $3$ phalanges, and an exoskeleton can be designed to interact with various numbers of phalanges to transmit actuator forces and to rotate finger joints. The number of interactions mostly depends on finger mobility. One approach to design finger components is to choose the same number of interaction points as the number of DoF. In other words, an exoskeleton can be designed with $4~DoF$ and $3$ interaction points~\cite{Fu2007, Wege2005, Lee2013, Weiss2013, Cempini2013, Chiri2009, Li2011, Wang2009, Polotto2012, Agarwal2013}, $3~DoF$ and $3$ interaction points (Figure~\ref{fig:3cp})~\cite{In2010, Connelly2010, Delph2013, Li2017, Popov2017, Mulas2005, Kobayashi2012, Kobayashi2013, Jo2014, Arata2013, Yap2015, Ryu2008, Jones2010, Hasegawa2008, DiCicco2004, Rahman2012, Allotta2015, Geo2016, Lee2015, Yamaura2009, Burton2011, Wei2017, Kim2017, Agarwal2015, Troncossi2012, Lince2017, Hocaoglu2009, Maeder-York2014}, $2~DoF$ and $2$ interaction points (Figure~\ref{fig:2cp})~\cite{Aubin2013, Decker2017, Jo2017, Taheri2014, Tong2010, Tang2011, Lu2016, Wu2010, Wei2013, Sarac2016} or $1~DoF$ and $1$ interaction point (Figure~\ref{fig:1cp_1})~\cite{Brokaw2011, Fiorilla2008}.

Devices with multiple interactions enhance the grasping stability during assistive and rehabilitation, and improve the haptic perception. Furthermore, they improve patients' safety by strictly limiting the spasticity. However, they might suffer from the design complexity of choosing high finger mobility. Mechanisms with $2$ interaction points can achieve $3~DoF$~\cite{Cui2015, Cortese2015} or $4~DoF$~\cite{Fontana2009} finger mobility. Alternatively, fingertip devices can achieve $2~DoF$~\cite{Gasser2015}, $3~DoF$ (Figure~\ref{fig:1cp_3})~\cite{Polygerinos2015, Gosselin2005, BenTzvi2015, Fang2009} or $4~DoF$~\cite{Bouzit2002, Iqbal2011, Sun2009, Sarakoglou2016, Ferguson2018} mobility. Even though having less number of interaction points simplifies the device mechanically, they might fail to reflect realistic interactions for certain haptic or assistive applications. For instance, a fingertip device can allow users to interact with objects and apply event-based forces, but cannot apply grasping forces on finger phalanges realistically.

Even though having less number of interaction points have simpler design and are easier to be worn, a generic exoskeleton should be designed with the same number of interaction points as the number of DoF.

\vspace*{-0.3\baselineskip}
\subsection{Mechanical Design}

\begin{figure}[h!]
  \centering
  \vspace*{-.75\baselineskip}
  \resizebox{3.2in}{!}{\includegraphics{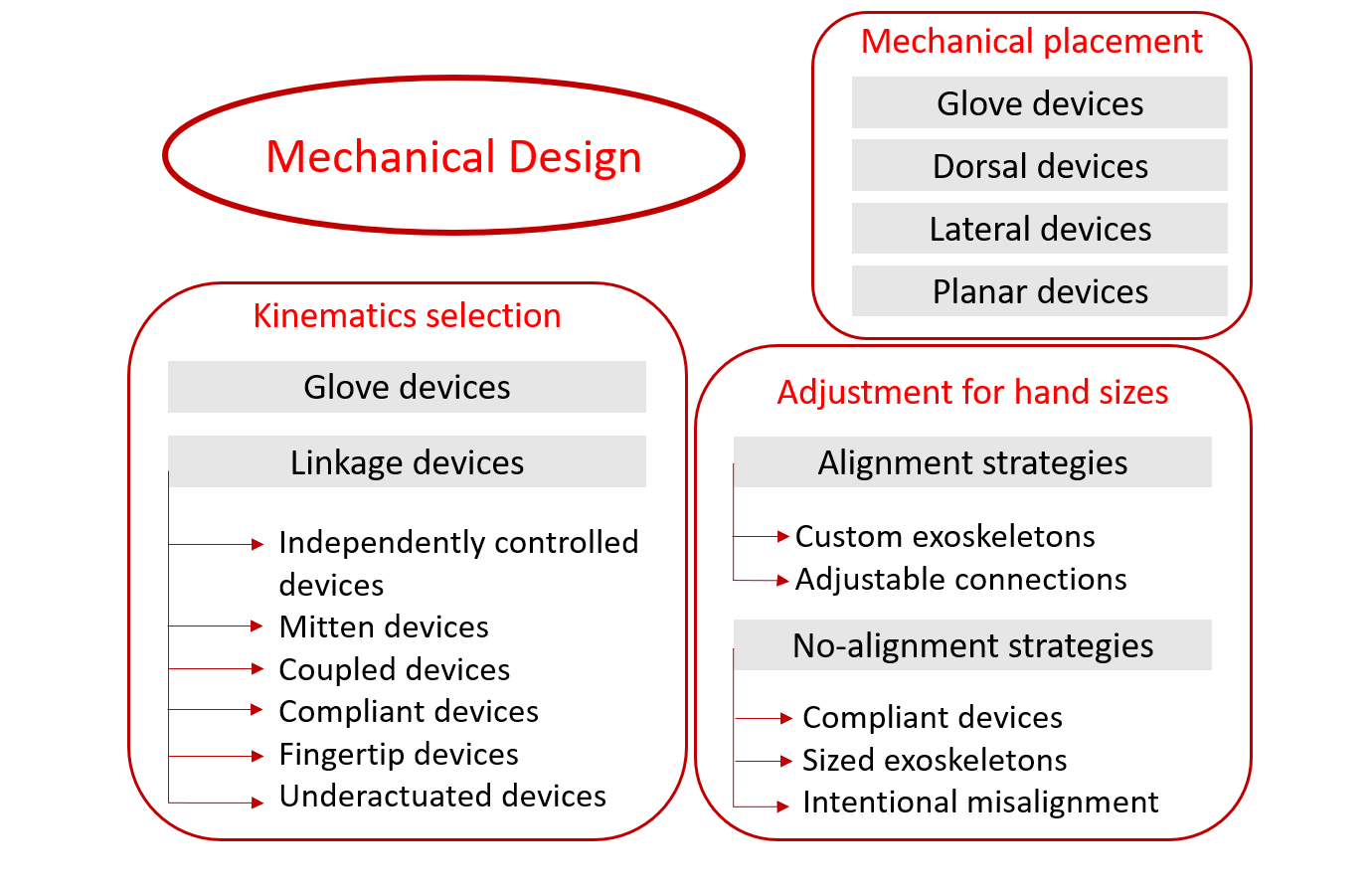}}
  \vspace*{-1\baselineskip}
  \caption{Possible design choices for mechanical design based on kinematics selection, mechanical placement and adjustment strategies for hand sizes.}
  \label{fig:mechanical}
  \vspace*{-1.2\baselineskip}
\end{figure}

Towards creating a hand exoskeleton, the next step of the designer should be how to achieve the mobility decisions through mechanical design. The mechanical design aspect can be handled based on kinematics selection, mechanical placement, and adjustment strategies for different hand sizes (see Figure~\ref{fig:mechanical}).

\begin{figure*}[t!]
\centering
\subfigure[Glove~\cite{Connelly2010} \label{fig:glove}]%
{\includegraphics[width=0.175\textwidth]{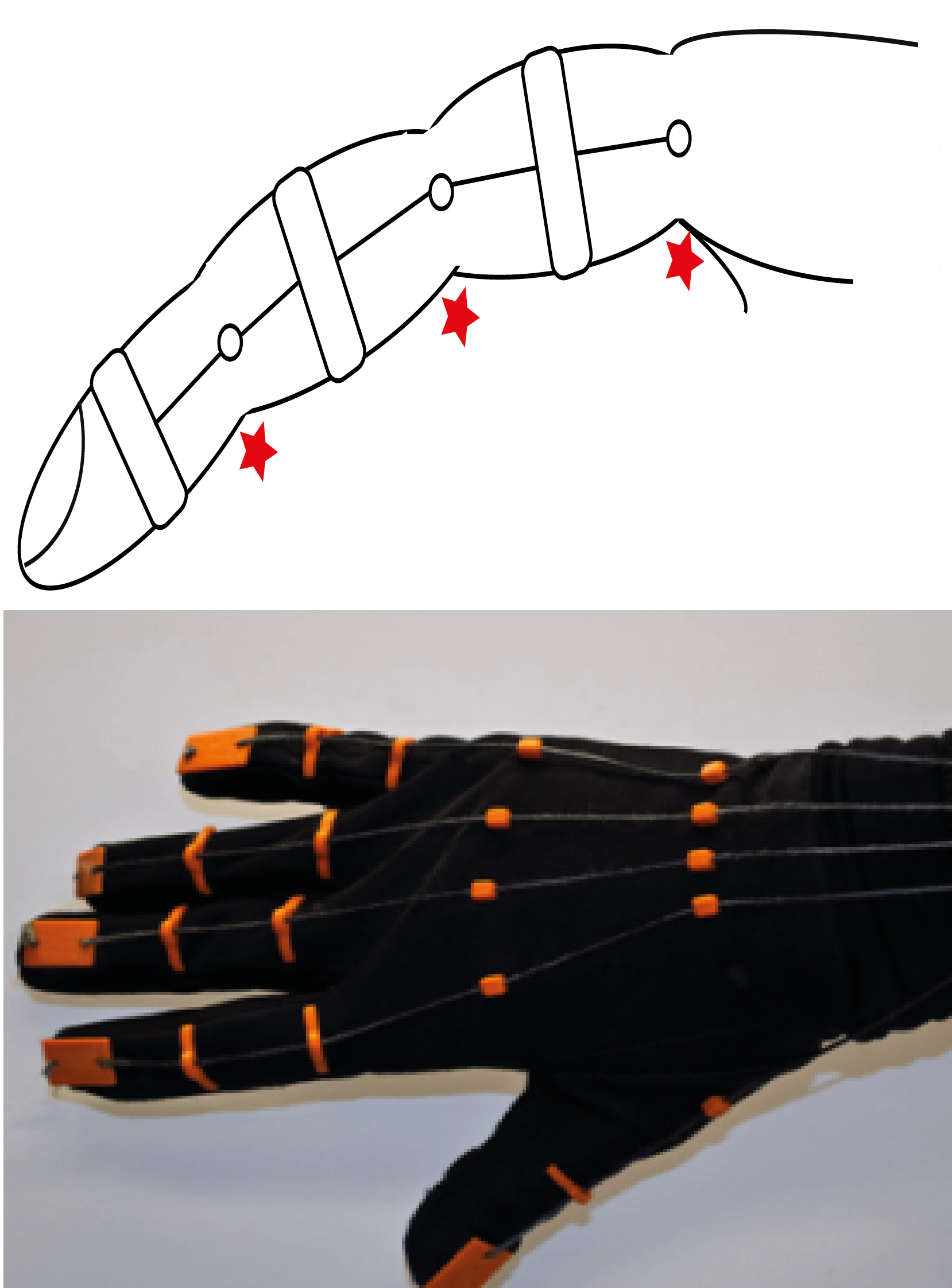}} \hspace{0.07\textwidth}
\subfigure[Glove with links~\cite{Jo2014} \label{fig:halfglove}]%
{\includegraphics[width=0.175\textwidth]{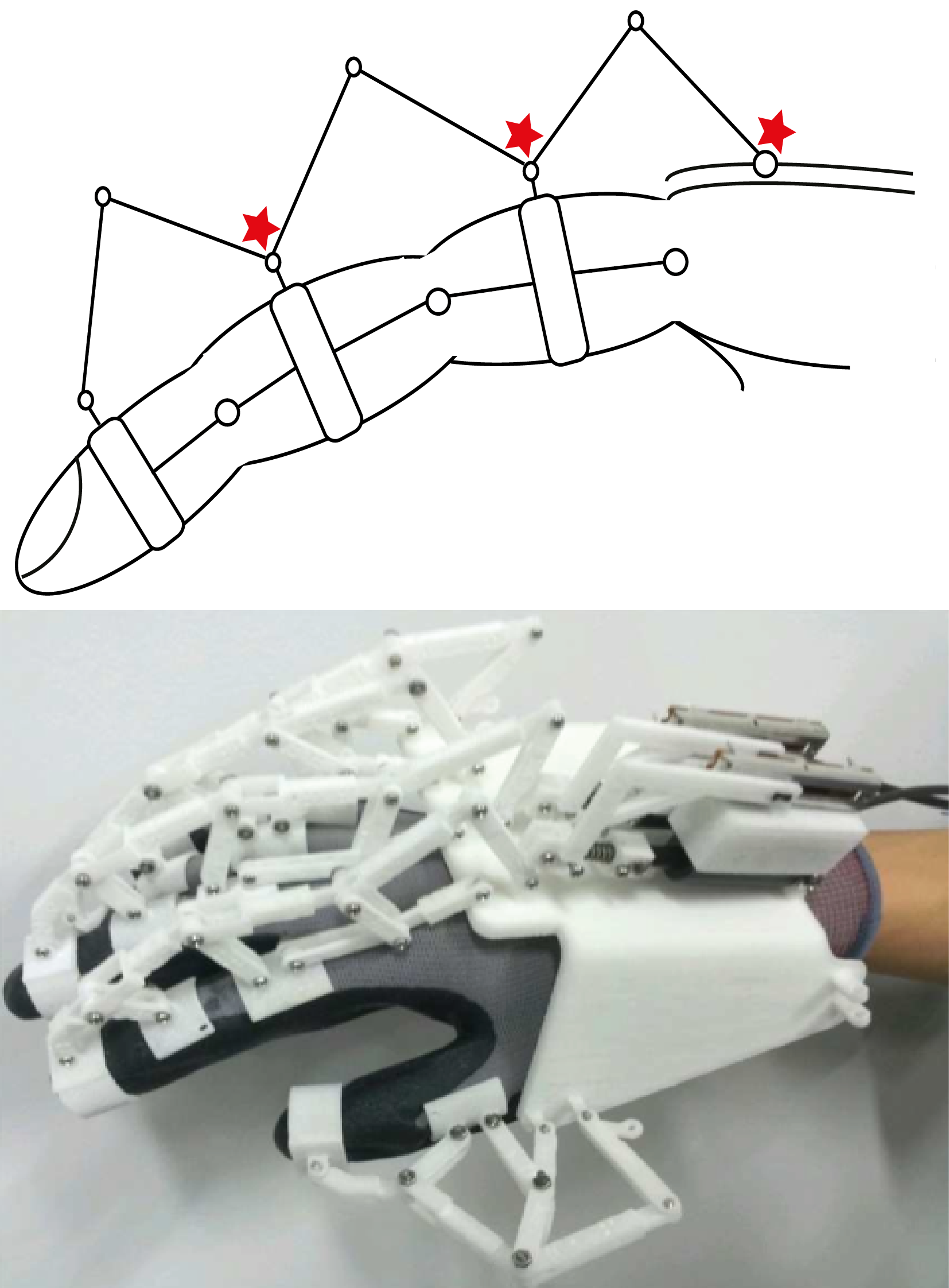}} \hspace{0.07\textwidth}
\subfigure[Indep. control~\cite{Agarwal2013} \label{fig:indep}]%
{\includegraphics[width=0.175\textwidth]{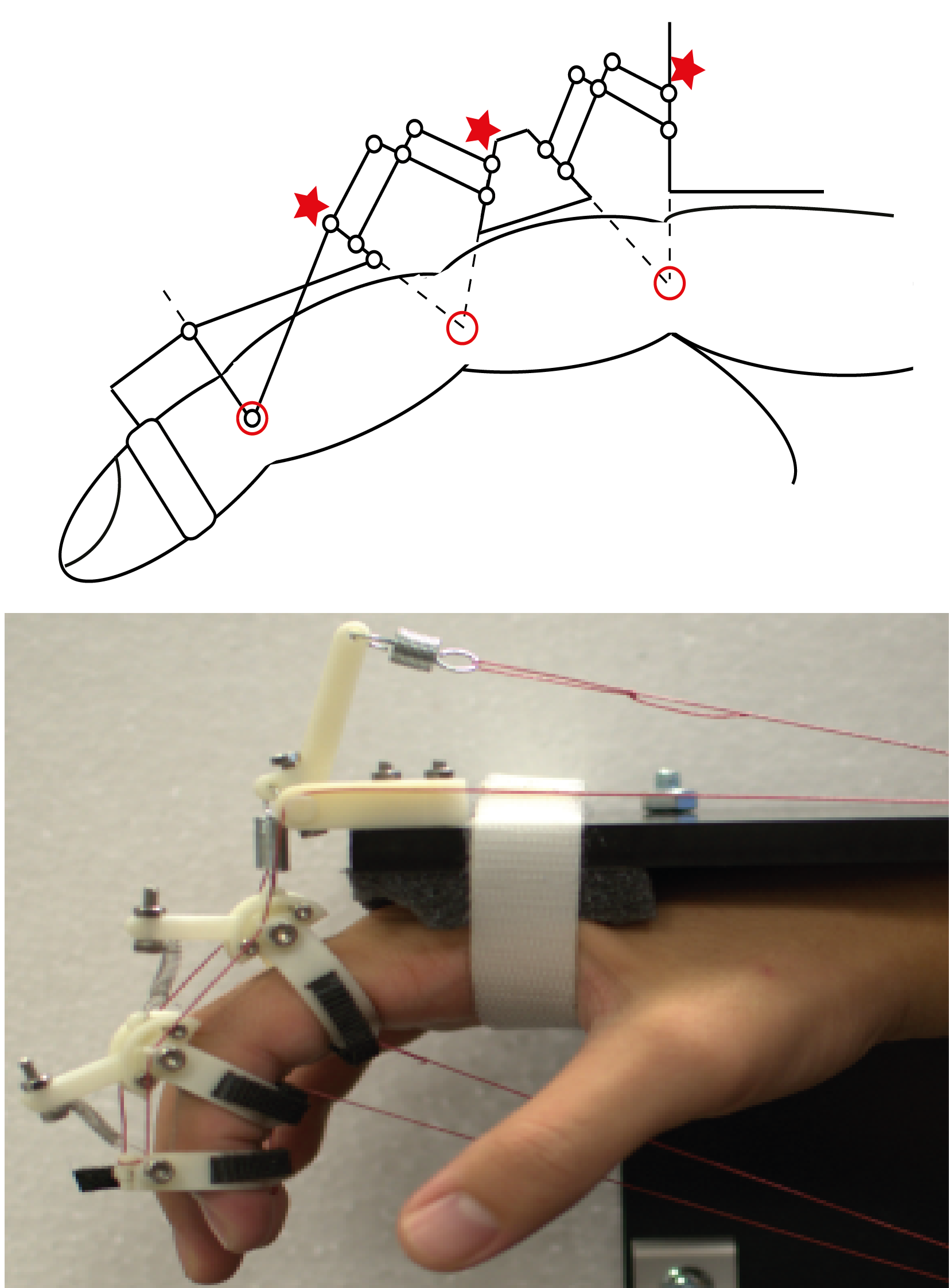}} \\ [-0.5em]
\subfigure[Fingertip device~\cite{Sun2009} \label{fig:fingertip_coupled}]%
{\includegraphics[width=0.175\textwidth]{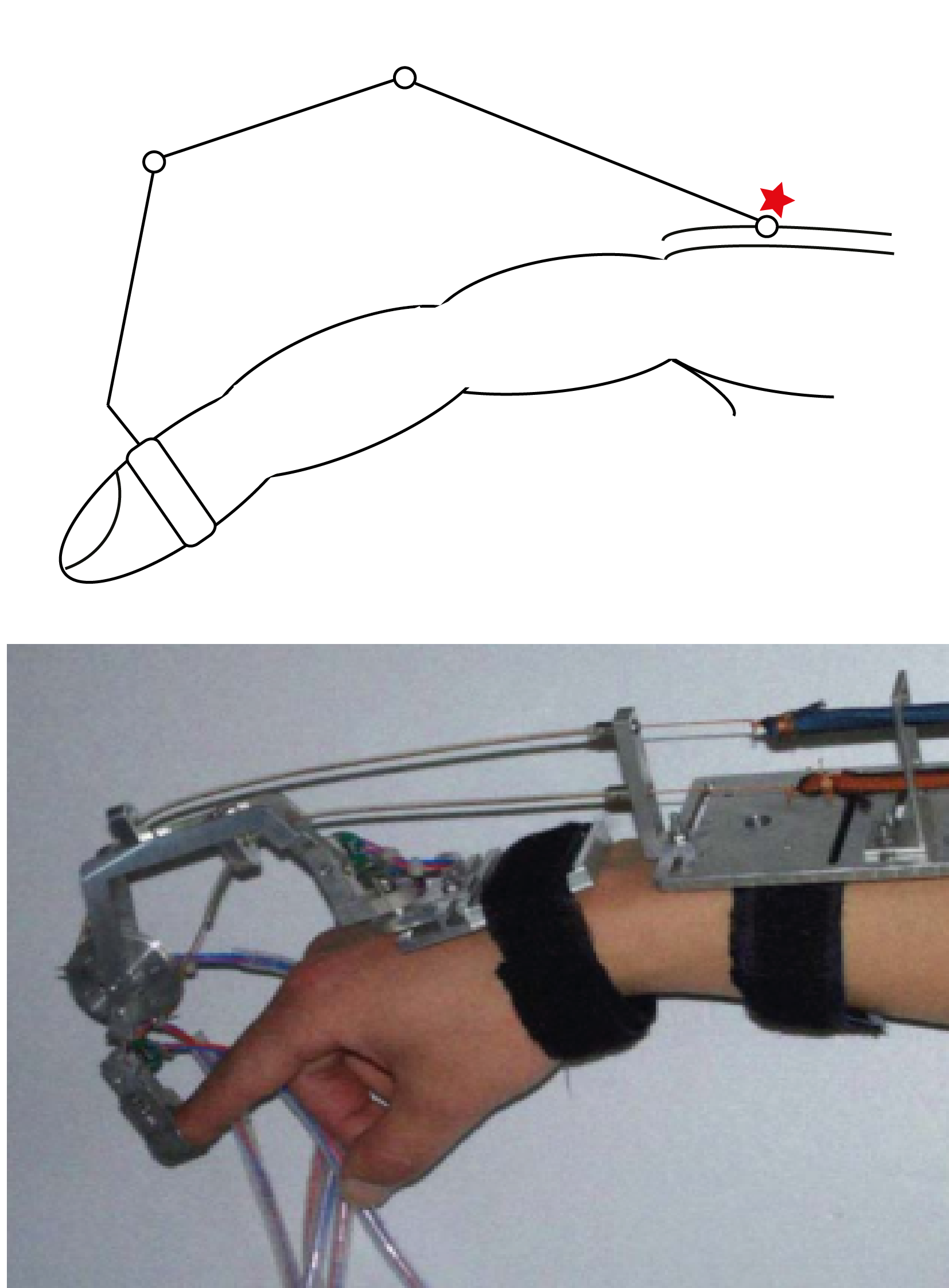}} \hspace{0.07\textwidth}
\subfigure[Coupled device~\cite{Troncossi2012} \label{fig:coupled}]%
{\includegraphics[width=0.175\textwidth]{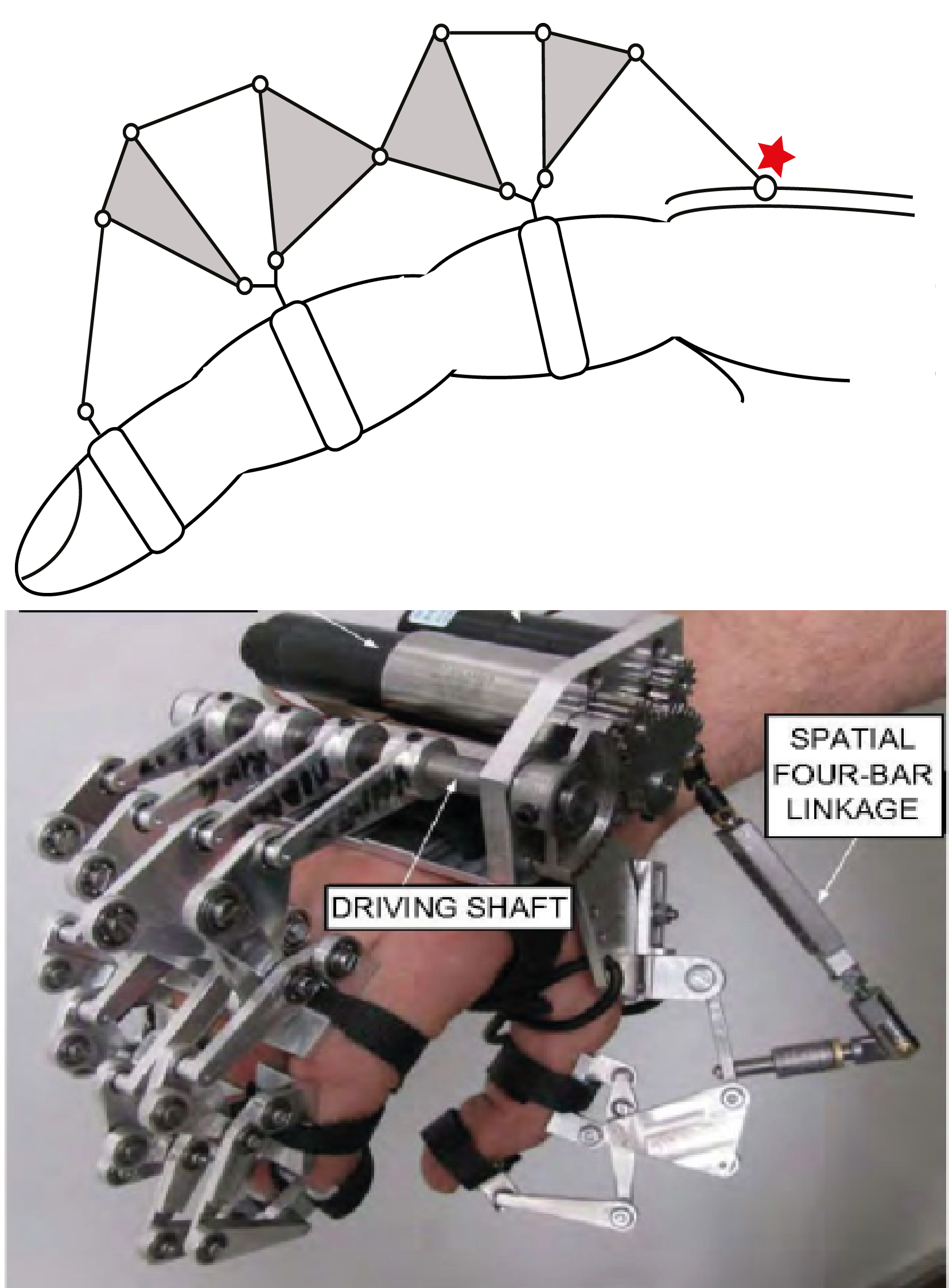}} \hspace{0.07\textwidth}
\subfigure[Underactuation~\cite{Ertas2014} \label{fig:underactuated}]%
{\includegraphics[width=0.175\textwidth]{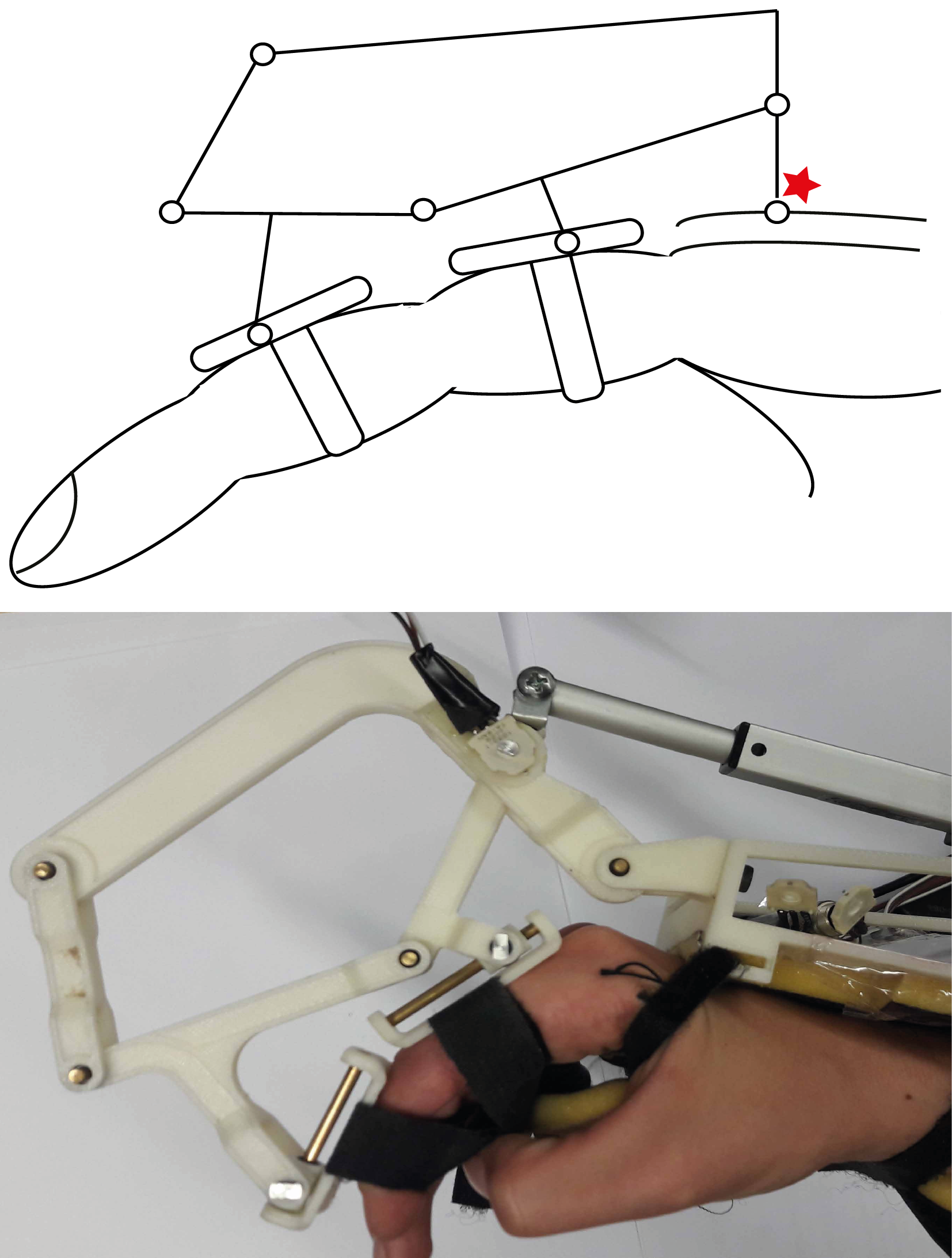}}
\vspace*{-.75\baselineskip}
\caption{Types of kinematics selections as hand exoskeletons: the black circles show the mechanical joints, while the red stars represent the actuated ones. Glove-based devices can track finger pose easily and efficiently but are hard to be worn. Linkage-based devices are lightweight, portable and easily wearable. Linkage-based devices can be categorized based on the finger mobility choices detailed in Section~\ref{sec:q3q4}.}
	\label{fig:kinematics}
\vspace*{-.75\baselineskip}
\end{figure*}

\subsubsection{Kinematics selection} \label{sec:q6}

The kinematics structure of a hand exoskeleton can be handled as glove-based or linkage-based devices. \textbf{Glove-based devices} require the user to wear a flexible glove equipped with sensors for motion tracking, and are perfect for haptic applications. They can assist/resist user's activity through cable transmission (Figure~\ref{fig:glove})~\cite{In2010, Connelly2010, Delph2013, Li2017, Popov2017}, or linkage transmission (Figure~\ref{fig:halfglove})~\cite{Mulas2005, Kobayashi2012, Kobayashi2013, Jo2014, Lee2013, Aubin2013, Decker2017, Jo2017}. 
Even though their wearability can be improved using Velcro connections in the palm~\cite{Decker2017} or half gloves~\cite{Mulas2005, Jo2017}, patients still have to reach an initial pose to wear the glove.

\textbf{Linkage-based devices} use mechanical links to form the finger components, and can be further categorized with independent joint control, MCP rotation only, full coupling, partial coupling, mitten style, fingertip connection, compliance and contact based underactuation. Devices with independent control have an individual actuator for each assisted finger joint (Figure~\ref{fig:indep})~\cite{Wang2009, Jones2010, Li2011, Hasegawa2008, Polotto2012, Taheri2014, Agarwal2013, Fiorilla2008, Brokaw2011}. These actuators are mostly placed remotely and their forces are transmitted through cables. Even though they can achieve full mobility, increasing the number of actuators significantly affects their cost and portability.


Linkage-based devices can be simplified in terms of the number of actuators with different kinematical structures. Mitten devices open/close the hand in a unique, repetitive way by coupling index, middle, ring and little fingers physically~\cite{Wei2013, Wu2010, Gasser2015, Lince2017, Troncossi2012}. Controlling the hand with $1$ or $2$ actuators simplifies the design and decreases the overall cost, but limit the mobility and task adjustability.

Coupled devices interact with user's finger from multiple points and move finger joints together with a ratio adjusted by mechanical links or differential system (Figure~\ref{fig:coupled}). Such mechanisms can control finger movements with $1$ actuator\cite{Cempini2013, Weiss2013, DiCicco2004, Fu2007, Wege2005, Tong2010, Chiri2009, Rahman2012, Tang2011, Allotta2015, Cui2015, Geo2016, Lee2015, Lu2016} or $2$ actuators~\cite{Yamaura2009, Burton2011, Fontana2009, Wei2017, Cortese2015, Kim2017, Agarwal2015, Ferguson2018}. Compliant devices couple finger joints through compliant elements~\cite{Arata2013}, artificial muscles~\cite{Ryu2008} or soft actuators~\cite{Polygerinos2015, Yap2015, Maeder-York2014} instead of rigid links. Their coupling ratio is set by the mechanical stiffness of these soft elements. They are low-cost, but suffer from mandatory mechanical adjustments to change the finger synergies.


Unlike coupled devices, fingertip devices interact with user's finger from a single point and control the fingertip position regardless how finger joints move (Figure~\ref{fig:fingertip_coupled})~\cite{Iqbal2011, Gosselin2005, Bouzit2002, BenTzvi2015, Fang2009, Sun2009, Sarakoglou2016}. Each finger component is controlled using a single actuator, so they are low-cost, easily wearable and portable. Not having strict mechanical connections around every finger phalange allows users to adjust tasks within the limits of their abilities. However, they cannot impose strict finger synergies, limit spastic movements for patients with disabilities or convey realistic information about virtual interactions.


Finally, underactuated devices based on contact forces control multiple finger joints with a single actuator by adjusting forces acting on finger phalanges automatically based on interaction forces, thanks to passive elements along the mechanism (Figure~\ref{fig:underactuated})~\cite{Hocaoglu2009, Sarac2016}. Each finger component is controlled using a single actuator, so they are low-cost, lightweight and portable. Passive elements along the mechanism ensures the device to be worn easily. Even though the actuator does not control the joints implicitly, alternative control strategies can improve the trajectory following tasks because they have multiple interactions for each finger (see Section~\ref{sec:operation}).


The kinematics of a generic exoskeleton should be consistent with the desired finger mobility. Full finger mobility can be achieved with linkage-based devices with independent control. Alternatively, finger joints can be coupled with underactuated linkage-based devices based on contact forces. Doing so, a single actuator controls a single finger component while adjusting the operation for different tasks automatically.


\begin{figure*}[h!]
\centering
\subfigure[Palmar device~\cite{Bouzit2002} \label{fig:palmar}]%
{\includegraphics[width=0.2\textwidth]{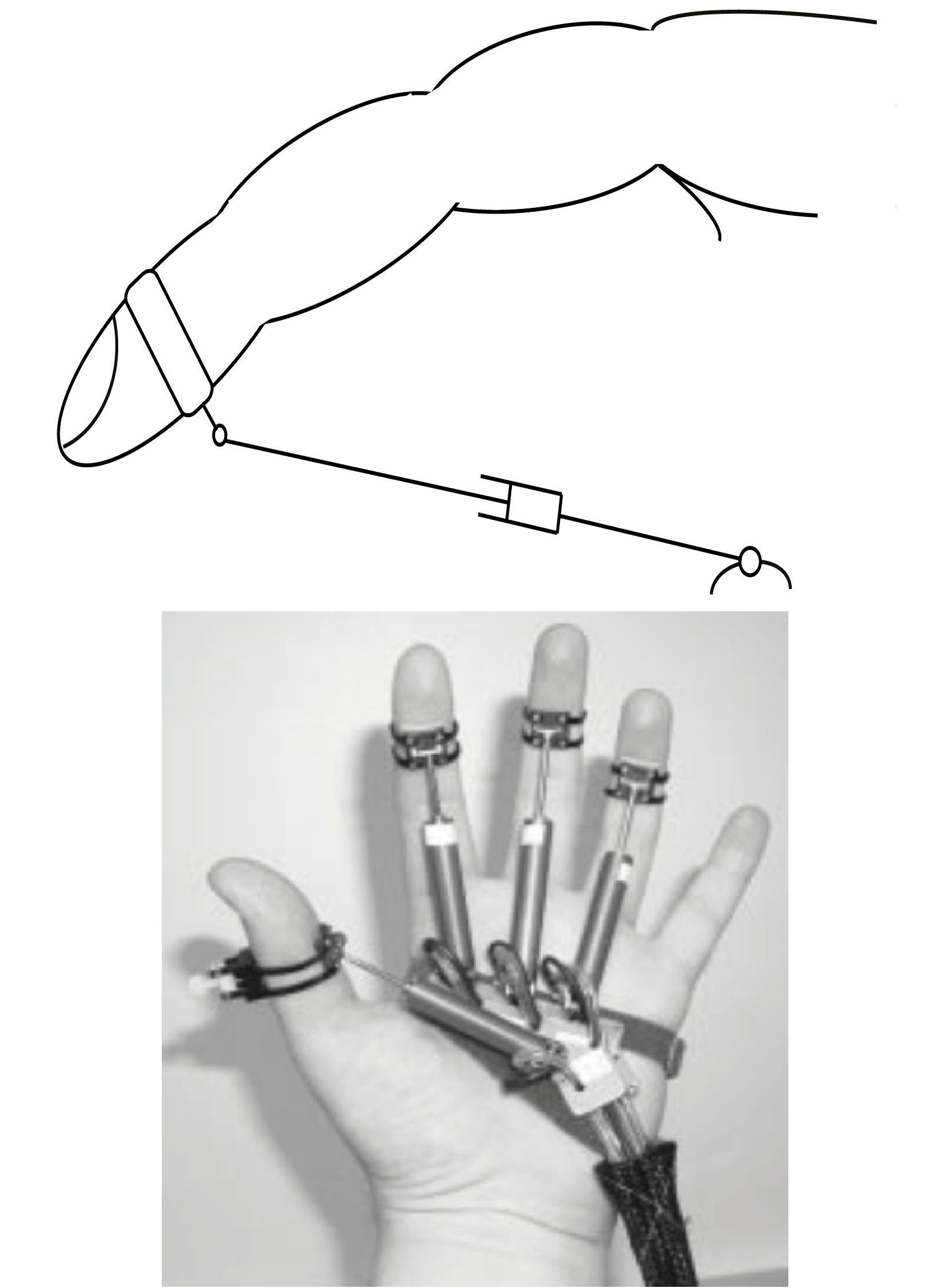}} \hspace{0.04\textwidth}
\subfigure[Lateral device~\cite{Hasegawa2011} \label{fig:lateral}]
{\includegraphics[width=0.2\textwidth]{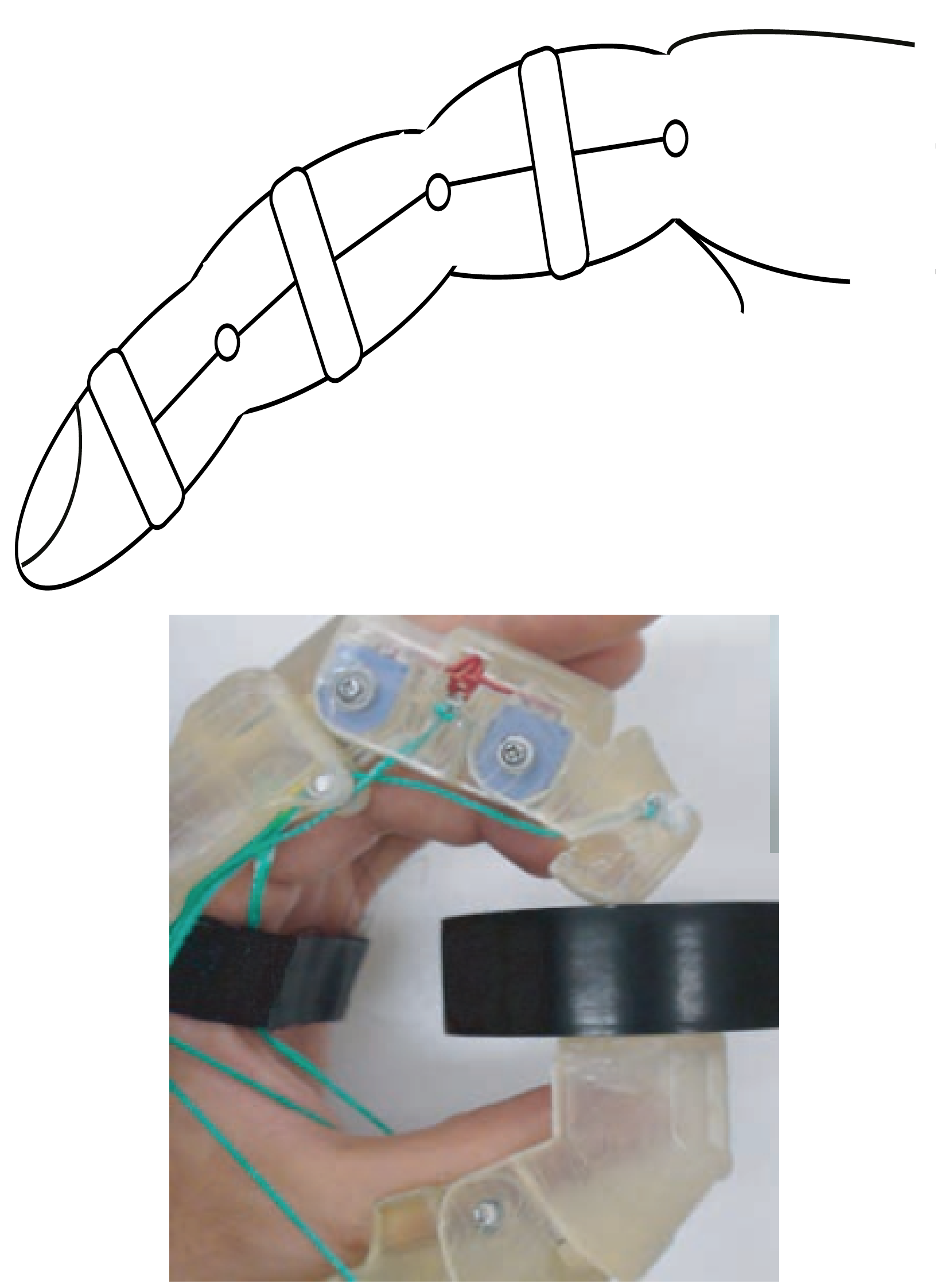}} \hspace{0.04\textwidth}
\subfigure[Dorsal device~\cite{Iqbal2015} \label{fig:dorsal}]%
{\includegraphics[width=0.2\textwidth]{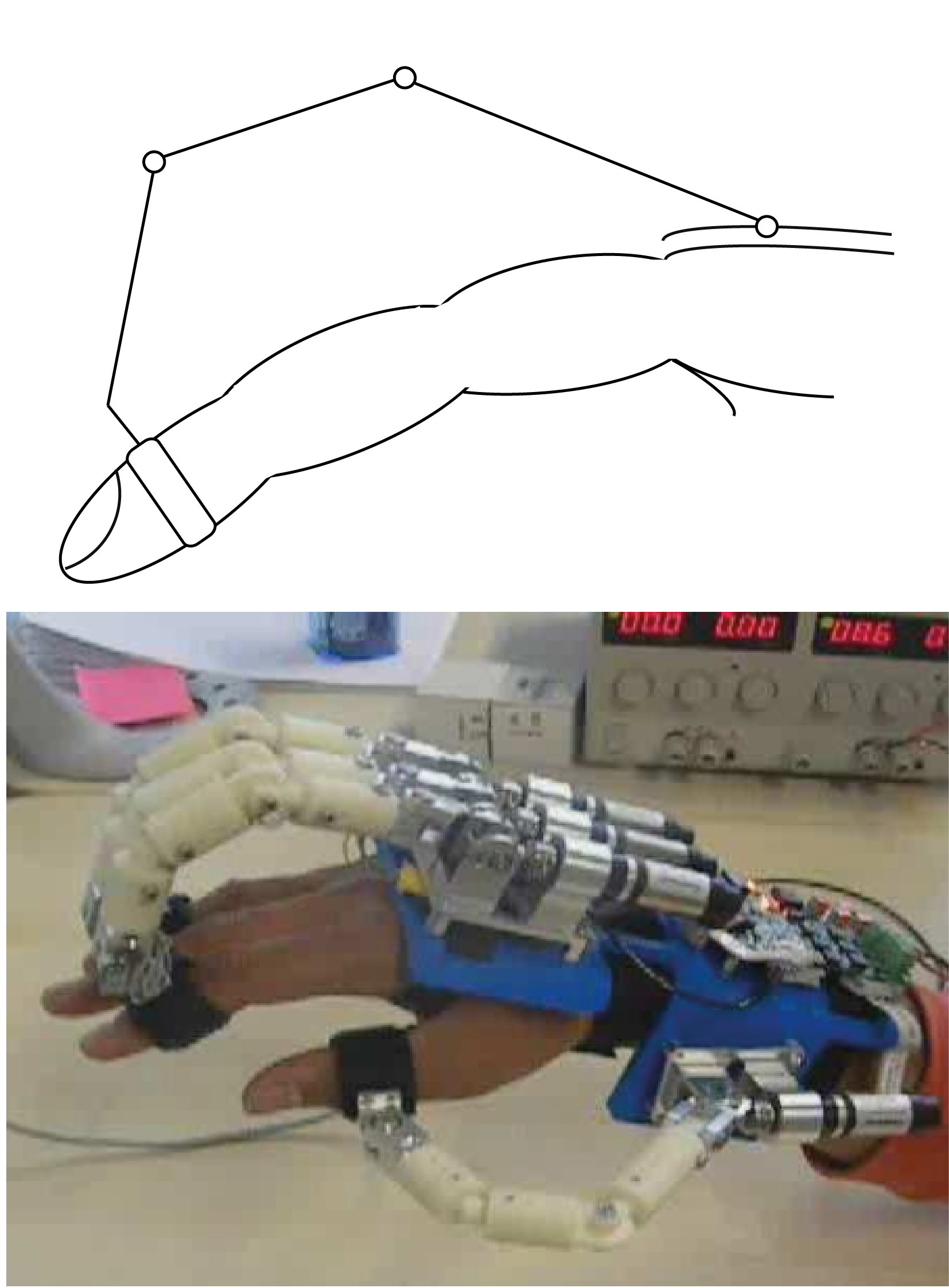}}
\vspace*{-.5\baselineskip}
\caption{Hand exoskeletons can be designed with different kinematics selections. }
\label{fig:placement}
\vspace*{-.75\baselineskip}
\end{figure*}

\vspace*{-0.5\baselineskip}
\subsubsection{Strategies for adjusting to different hand sizes} \label{sec:q7}

The society has a wide range of hand sizes~\cite{Buryanov2010}, and a hand exoskeleton should operate correctly and comfortably for all users~\cite{Morel2012}. Exoskeletons with a single interaction point~\cite{Gosselin2005, Sarakoglou2016, Fiorilla2008} can neglect such variety, since they control the fingertip pose without imposing strict trajectory for finger joints. For exoskeletons with multiple interaction points, several adjustment strategies can be found in the literature:

\vspace*{0.3\baselineskip}
\textbf{Alignment strategies} require mechanical and finger joints to be aligned, such that the exoskeleton can fit on user's hand accurately, and actuator forces can be mapped into perceived ones directly. The first alignment strategy is to manufacture a custom exoskeleton for each user individually~\cite{Delph2013, Hasegawa2008, Burton2011, Weiss2013, Cui2015}. A custom exoskeleton must be designed with variable link lengths corresponding to user's hand size. Such an exoskeleton must be manufactured individually, so the user must agree to purchase it for personal use. Due to the lack of mass production, the overall cost of the device is expected to be high. Even though this strategy might be suitable for assistive or haptic applications, it is not applicable for clinical use, where a single device is expected to serve for multiple patients in a day.

Alternatively, an exoskeleton can align mechanical and finger joints through adjusted mechanical connections and links~\cite{Popov2017, Aubin2013, Decker2017, Brokaw2011, Bouzit2002, Iqbal2011, DiCicco2004, Fang2009, Wang2009, Jones2010, Li2011, Polotto2012, Taheri2014, Fu2007, Wege2005, Tong2010, Geo2016, Agarwal2013, Wei2017, Agarwal2015, Wei2013}. The user wears the device before operation and a technician fixes a slider-screw system for fitting. Even though it requires a crucial preparation process, the exoskeleton can fit all users in the end. The constant need for a technician's presence might make such an exoskeleton suitable for clinical settings more than home therapy.

\vspace*{0.3\baselineskip}
\textbf{No-Alignment Strategies} accept the misalignments between mechanical and finger joints, and address the issue of hand sizes in other ways. Increasing the compliance of the actuator~\cite{In2010, Connelly2010, Li2017, Polygerinos2015, Yap2015, Ryu2008, Wu2010} or the mechanical links~\cite{Jo2014} transmits lower interaction forces, hence minimizes the after effects of misalignments. However, the output forces might be insufficient for certain rehabilitation or assistive applications.

A hand exoskeleton can be designed in small, medium and large sizes, such that the misalignment between mechanical and finger joints can be limited~\cite{Mulas2005, Kobayashi2012, Lee2013, Jo2017, Arata2013, Sun2009, Gasser2015, Rahman2012, Lu2016}. Even though misalignments are not prevented, they are ensured not to harm users. Finally, a designer can place passive joints along the mechanical structure to turn additional loads, which are caused by misalignments, into motion~\cite{BenTzvi2015, Sarac2016, Ertas2014, Chiri2009, Cempini2013, Tang2011, Allotta2015, Lee2015, Fontana2009, Yamaura2009, Cortese2015, Kim2017, Troncossi2012, Lince2017, Ferguson2018}. Such an exoskeleton adapts its behavior for different hand sizes automatically. Designing sized exoskeletons and passive joints are the best practices for generic exoskeletons, thanks to their usability and preparation time. Furthermore, since they can be mass produced, they can be low-cost.

\subsubsection{Mechanism Placement} \label{sec:q8}

Finally, the designer should device where to place the finger components with respect to the fingers. This design selection is especially important for linkage-based exoskeletons, such that transmission units can be placed on dorsal, lateral or palmar side of fingers (see Figure~\ref{fig:placement}).

\vspace*{0.3\baselineskip}
\textbf{Palmar devices} consist of mechanical or transmission components placed inside the palm of the hand (see Figure~\ref{fig:palmar})~\cite{Bouzit2002, Connelly2010, Delph2013, Lee2013}. Unfortunately, they prevent users to get in touch with real objects for assistive use.

\vspace*{0.3\baselineskip}
\textbf{Lateral devices} consist of mechanical or transmission components placed on both sides of finger phalanges (see Figure~\ref{fig:lateral})~\cite{Jones2010, Hasegawa2008, Polotto2012, Weiss2013, Cempini2013, Cortese2015, Wei2013, Gasser2015, Mulas2005}. Finger joints can be rotated independently through cable transmission or remote center of motion (RCM). These devices free the palm of the hand for future interactions in the real environment. However, they might suffer from possible collisions for multi-finger implementations, especially when abduction/adduction of MCP is allowed. Compared to other options, lateral devices might be harder to be worn by patients with disabilities, so their use for rehabilitation or assistive should be reconsidered.

\vspace*{0.3\baselineskip}
\textbf{Dorsal devices} consist of mechanical or transmission components placed on top of the finger phalanges (Figure~\ref{fig:dorsal})~\cite{Sarac2016, Wu2010, Troncossi2012, Lince2017, Fontana2009, Yamaura2009, Burton2011, Agarwal2013, Wei2017, Kim2017, Agarwal2015, DiCicco2004, Fu2007, Wege2005, Tong2010, Chiri2009, Rahman2012, Tang2011, Allotta2015, Cui2015, Geo2016, Lee2015, Lu2016, Wang2009, Li2011, Taheri2014, Gosselin2005, Iqbal2011, BenTzvi2015, Fang2009, Sun2009, Sarakoglou2016, Arata2013, Yap2015, Brokaw2011, Fiorilla2008, Li2017, Kobayashi2012, Kobayashi2013, Jo2014, Aubin2013, Decker2017, Jo2017, Cempini2013, Weiss2013, Lee2013, Connelly2010, Delph2013, Polygerinos2015, Hasegawa2008, Mulas2005, Popov2017, Gasser2015, In2010, Jones2010, Polotto2012, Maeder-York2014, Ferguson2018}. Doing so, the collision between multiple finger components can be minimized while user's palm is free for future interactions with real objects. They do not possess any strong limitation regarding the number of finger components to be manufactured or the performance, and can be used for all possible target applications.

Regardless the placement of finger components, linkage-based exoskeletons are attached to user's fingers through rings or flexible attachments. Since there is no recorded impact of mechanism placement on perception during finger opening/closing, we can assume that actuator forced can be distributed around finger phalanges naturally.

\vspace*{-0.3\baselineskip}
\subsection{Actuation}

\begin{figure}[t!]
  \centering
  \resizebox{3.2in}{!}{\includegraphics{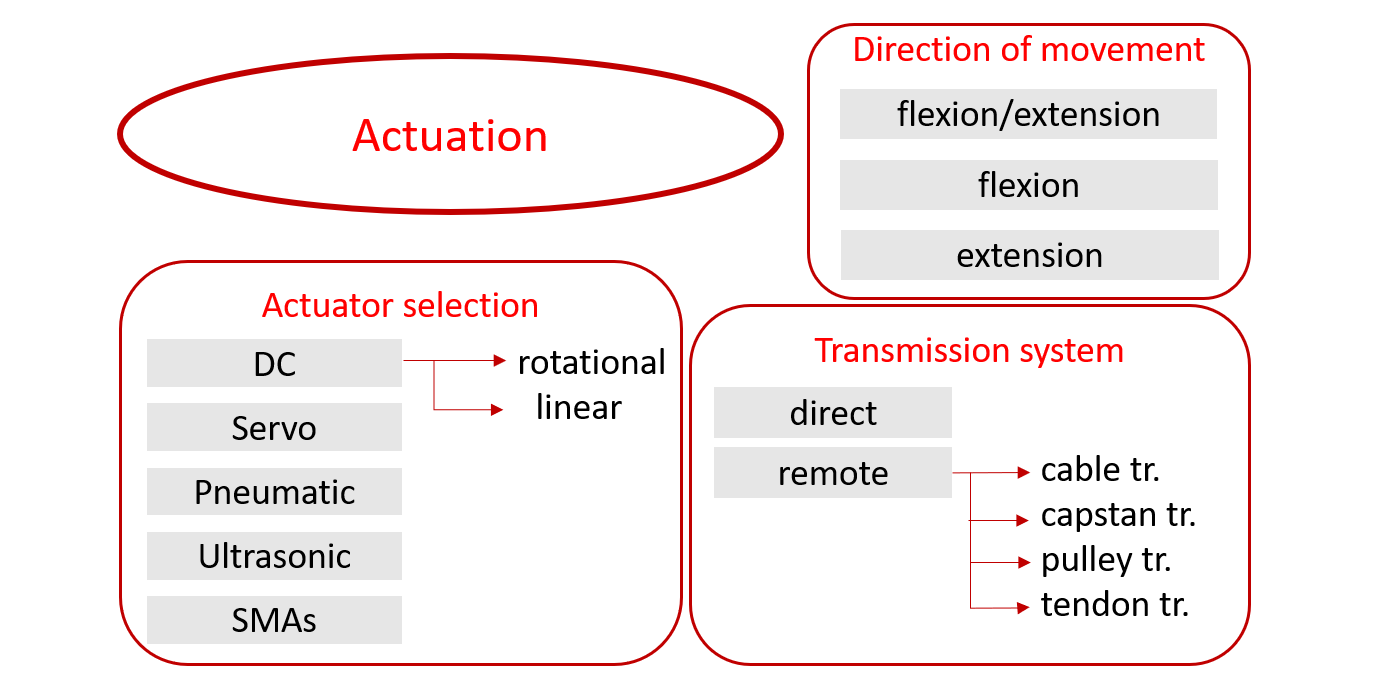}}\label{fig:actuation}
  \vspace*{-.5\baselineskip}
  \caption{Possible design choices for actuation technologies based on actuator selection, transmission system and direction of movement.}
  \vspace*{-1.25\baselineskip}
\end{figure}

An exoskeleton can assist/resist user's fingers through actuator and transmission technologies. In this section, we will investigate the exoskeletons in the literature based on actuator selection, direction of movement and transmission system from the perspective of achieving generic exoskeletons (see Figure~\ref{fig:actuation}).

\subsubsection{Actuator type} \label{sec:q10}

Even though there are some exceptional studies that apply assistance based on wrist activity~\cite{Bortoletto2017} or resistance using springs~\cite{Brokaw2011}, most of the exoskeletons move user's fingers through active manipulation. Such a manipulation can be achieved through different actuator types.

\vspace*{0.3\baselineskip}
\textbf{DC motors} are the most popular technology since they are highly available in the market, reliable and easily controllable. Linear movement can be achieved using linear DC motors~\cite{Jo2017, Arata2013, Taheri2014, Tong2010, Rahman2012, Cui2015, Sarac2016, Jo2014} or rotational DC motors with linear sliders~\cite{Chiri2009, Cortese2015}. Then, rotational movement can be achieved using brushed motors~\cite{Popov2017, Decker2017, Lee2013, Fiorilla2008, Gosselin2005, Iqbal2011, BenTzvi2015, Sarakoglou2016, Wang2009, Jones2010, Li2011, Polotto2012, Fu2007, Wege2005, Weiss2013, Cempini2013, Geo2016, Lee2015, Lu2016, Fontana2009, Wei2017, Troncossi2012, Kim2017, Hasegawa2008, Hocaoglu2009, Popov2017, Ferguson2018} or brushless motors~\cite{Fang2009, Wei2013, Gasser2015, Lince2017, In2010}. Linear motors are simpler to be placed on top of the hand for coupled finger opening/closing, while rotational motors are mostly backdriveable and provide unlimited movement. Furthermore, brushed motors have low-cost, simple wiring, compact design and easy control but require maintenance, cause vibration and lose torque in high speeds due to friction.

\vspace*{0.3\baselineskip}
\textbf{Servo motors} can be defined as rotational DC motors with a limited workspace~\cite{Delph2013, Mulas2005, Aubin2013, Yamaura2009, Allotta2015}. They are fast, and can achieve high output torque and accurate position control; but require a special driving circuit for control and have higher cost compared to DC motors. \textbf{Ultrasonic motors (USMs)} can also be defined as rotational DC motors powered by ultrasonic vibration~\cite{Tang2011}. They are silent, light weight and efficient in terms of output force, but they suffer from hysteresis and temperature increase over time.

\vspace*{0.3\baselineskip}
\textbf{Pneumatic actuators} control the hydraulic or air flow through compressors, using pneumatic cylinders~\cite{Bouzit2002, DiCicco2004, Burton2011}, air balloons~\cite{Li2017}, hydraulic pump~\cite{Ryu2008}, air bladder~\cite{Connelly2010}, flexible thermoplastic fabrics~\cite{Yap2015}, soft actuation~\cite{Polygerinos2015} or pneumatic artificial muscles~\cite{Sun2009, Agarwal2013, Wu2010, Maeder-York2014}. They can achieve high, adjustable force and speed at low-cost. The size of the compressor and its storage lead the exoskeletons to be controlled remotely. Even though pneumatic actuators are not necessarily compliant, they consequently increase the overall compliance as mentioned in Section~\ref{sec:q7}.

\vspace*{0.3\baselineskip}
\textbf{Shape memory alloy actuators (SMAs) }use deformation of materials upon heating and cooling at critical temperatures~\cite{Kobayashi2012, Kobayashi2013}. Even though they have high power-to-weight ratio, their output motion is hysteresis, highly nonlinear and saturated. As a result, their control is challenging~\cite{Kumar2008}.

Actuation types do not possess strong limitation about applications or tasks. Therefore, any actuator type can be selected for a generic exoskeleton as long as they are low-cost, easily controllable and effective in terms of output forces.

\subsubsection{Transmission units} \label{sec:q9}

The actuators should be connected to the mechanical structure through alternative transmission strategies. The simplest transmission scenario is designing a direct-drive system, such that the actuators are placed on top of the hand or along the mechanism, while the actuator shafts are attached to mechanical components directly~\cite{Sarac2016, Hocaoglu2009, Iqbal2011, Troncossi2012, Arata2013}. Even though direct-drive is preferable to improve the portability, the chosen actuators should be highly miniaturized and lightweight.

If the chosen actuators are big and heavy, they should be located away from the exoskeleton and their forces should be transmitted remotely through cables~\cite{Cempini2013, Wege2007, Delph2013, Arata2016, Chiri2009, Agarwal2015, Li2011, Wang2009, Fu2007, Kobayashi2012, Hasegawa2008, Lee2013, Li2011, Aubin2013, Jones2010, Polotto2012, Taheri2014, DiCicco2004, Wege2005, Weiss2013, Allotta2015, Lu2016, Yamaura2009, Agarwal2013, Ferguson2018}, capstan systems~\cite{Fontana2009, Ertas2014, Decker2017}, tendons~\cite{In2010, Popov2017, Cortese2015, Wei2013, Gasser2015}, or pulleys~\cite{Geo2016}. Even though choosing big actuators can create high output forces for the exoskeleton, remote transmission limit the workspace of users.

Both transmission strategies can be implemented for all application types. The designers should make the selection based on the actuator decision.

\subsubsection{Direction of movement} \label{sec:q11}

Even though the majority of actuators are bidirectional, certain rotational DC motors and pneumatic motors are not. If the chosen actuator is unidirectional, then the designer should decide how to use them for finger movements. One approach is to assist user's fingers in one direction actively, and to leave the other direction passive. The active assistance can be used either to open the finger for rehabilitation~\cite{Chiri2012, Brokaw2011, Agarwal2013, Connelly2010}, or to close the finger for assistive use~\cite{In2010, Lince2017, DiCicco2004}. Devices with active flexion cannot be used for haptic use due to the lack of resistive forces, while devices with active extension cannot be used for assistive use due to the lack of assistive forces. This is why leaving one direction passive cannot be chosen for a generic exoskeleton, even though they provide simple and effective solutions for specific tasks.

The second approach is to achieve bidirectional movement using multiple actuators and transmission units~\cite{Fu2007, Weiss2013, Cempini2013, Yamaura2009, Gasser2015, Wang2009, Jones2010, Li2011, Sun2009}. Bi-directional movements can be adopted for all target applications with no specific limitations. Achieving bi-directional movements might make exoskeletons bulkier and more expensive due to the increased number of actuators. Even though choosing bidirectional actuators is the best choice for generic exoskeletons, the designer should equip multiple actuators if unidirectional actuators are chosen for a specific purpose.


\vspace*{-0.3\baselineskip}
\subsection{Operation strategies} \label{sec:operation}

\begin{figure}[t!]
  \centering
  \resizebox{3.2in}{!}{\includegraphics{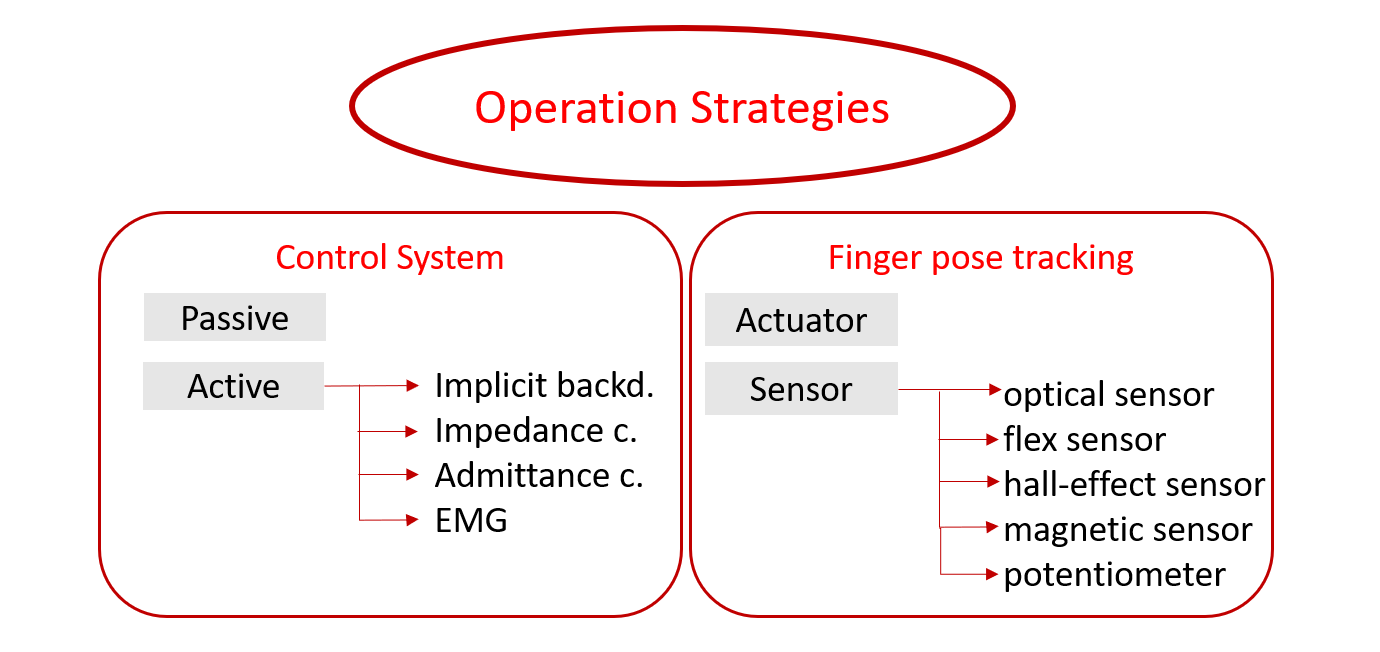}}
  \vspace*{-.5\baselineskip}
  \caption{Possible design choices for operational strategies based on control and finger pose tracking strategies.}
  \label{fig:operational}
  \vspace*{-1.25\baselineskip}
\end{figure}

The design of a hand exoskeleton is completed once the mechanical structure is equipped with actuators and transmission units. Then the designer should decide how to control and track user's fingers during operation (see Figure~\ref{fig:operational}).

\subsubsection{Control} \label{sec:q12}

Control strategies for existing hand exoskeletons can be categorized mainly as active and passive, based on how much users participate to the task~\cite{Jeong2013}.

\vspace*{0.3\baselineskip}
\textbf{Passive control strategies} control the exoskeleton to follow a strict trajectory or to reach a specific target. As the device leads their fingers, the user is asked to obey the movement. The control strategies can be designed based on position~\cite{Iqbal2011, Fu2007, Rahman2012, Connelly2010, Cui2015, Kim2017, Jo2017, Troncossi2012, Popov2017, Allotta2015, Weiss2013, Arata2013, Gasser2015, Wang2009, Sarac2016, Yamaura2009, Wege2005, Polygerinos2015, Yap2015, Burton2011, Li2017, Bouzit2002, Wu2010, Ryu2008, Cempini2015, Sun2009, Ferguson2018} or velocity~\cite{Jo2014, Cempini2015}. Even though passive exercises can be used to treat disabilities of patients during rehabilitation, they might cause patients to lose interest during long, intense therapy sessions. They can be used for assistive applications as long as they are triggered by an external state, such as an additional sensor or a condition satisfied by an arm exoskeleton. However, they are impractical for haptic use.

\vspace*{0.3\baselineskip}
\textbf{Active control strategies} control the exoskeleton to assist/resist user's fingers based on user's performance as they are in charge of following a trajectory or reaching a target. One way to achieve active control is to adopt implicit backdriveability, which requires actuation, transmission and mechanical units to be chosen accordingly. With implicit backdriveability, the user can move their fingers freely even if the exoskeleton is attached to their fingers with no control~\cite{Hocaoglu2009, Wang2009, Jones2010, Li2011, Polotto2012, Taheri2014, Chiri2009, Aubin2013, Tong2013, Decker2017, Connelly2010, Ryu2008, Arata2013}. The backdriveable devices can be controlled with passive strategies when the user fails to keep their performance within a predefined range.

Implicit backdriveability cannot be achieved if the mechanical and actuator components of the designed exoskeleton require high backdrive forces or cause high friction. If so, backdriveability can be achieved actively using force control techniques based on impedance~\cite{Wei2017, Agarwal2015} or admittance~\cite{BenTzvi2015, Sarac2018, Kobayashi2012, Kobayashi2013, Sarakoglou2016, In2015, Gosselin2005, Sun2009, Ferguson2018, Jo2014}. These techniques require additional force sensors to be included for the exoskeleton, such that user's intention to move can be measured and be used as a control reference for the exoskeleton. In either case, backdriveability can easily be used by all target applications and improve user's safety during operation.

Furthermore, user's intention can be detected through additional sensors, such as electromyography (EMG) sensors~\cite{Mulas2005, Delph2013, Arata2013, DiCicco2004, Wege2007, Tong2013, Troncossi2012, Lince2017} or active bioelectric potential electrodes~\cite{Hasegawa2011}. Then, these measurements are used to create a control reference for passive control strategies in an online manner. Similarly, bilateral teleoperation tasks can be developed by controlling the device passively to follow the reference set by user's other hand~\cite{Yamaura2009, Fu2007}. These assist-as-needed or bilateral control strategies are useful for rehabilitation or assistive applications but their use for haptics is out of context.

\subsubsection{Finger pose estimation}

A generic hand exoskeleton must track user's finger movements efficiently during operation. The exoskeletons in the literature adopt various strategies to track finger movements, mostly depending on mechanical and actuation choices.

\vspace*{0.3\baselineskip}
\textbf{Actuator displacements} reveal the finger pose directly with high quality for the exoskeletons with independent finger control~\cite{Brokaw2011, Wang2009, Li2011, Jones2010, Hasegawa2011, Polotto2012, Taheri2014, Aubin2013, Agarwal2013, Fiorilla2008}. Similarly, coupled exoskeletons with constant joint ratio between joint rotations track finger movements using the actuator displacements and this ratio~\cite{In2010, Mulas2005, Jo2017, Arata2013, Fang2009, DiCicco2004, Tong2010, Rahman2012, Tang2011, Cempini2013, Allotta2015, Geo2016, Lu2016, Fontana2009, Yamaura2009, Burton2011, Wei2017, Cortese2015, Kim2017, Troncossi2012, Gasser2015, Lince2017}. Using actuator displacements directly result in simple operational strategies and high quality tracking performance.

\vspace*{0.3\baselineskip}
\textbf{Additional sensors} are needed for exoskeletons with other kinematics selections, when the actuated joints are not directly mapped into finger joints. For glove-based exoskeletons, flex sensors are placed along user's finger joints to measure the finger pose directly~\cite{Wei2013, Popov2017, Connelly2010, Li2017, Kobayashi2012, Jo2014, Decker2017, Ryu2008}. Such flex sensors are low-cost, lightweight and of high quality. Since these flex sensors should be grounded along finger joints to measure the finger pose directly, they require a texture-based interface.

Furthermore, additional sensors can track user's finger movements when inserted along mechanical joints, which are aligned with finger joints. These sensors can be chosen among hall-effect sensors~\cite{Fu2007, Wege2005, Weiss2013, Fiorilla2008} or potentiometers~\cite{Lee2015}. The alignment between mechanical and finger joints measure the finger pose directly, so the measurements are quick and of high quality, while the sensors are mostly low-cost and lightweight. However, such direct pose tracking can be implemented only for exoskeletons with RCM mechanisms.

Non-contact optical~\cite{Lee2013, Kobayashi2012, Arata2013, Burton2011, Chiri2012} or magnetic sensors~\cite{Sun2009} require markers to be placed on finger phalanges or finger joints.  They can be applied only if the exoskeleton allows for these markers to be placed on user's fingers without optical interface. It is important to note that in case of interference, the continuity of finger pose might be disturbed.

The sensor implementations discussed above require certain kinematics decisions. If a hand exoskeleton does not satisfy any of these properties, forward kinematics computation can be used to estimate the finger pose using additional sensors attached along random mechanical joints. Hall-effect sensors~\cite{Bouzit2002, Wu2010, Wege2005}, optical encoders~\cite{Gosselin2005, BenTzvi2015}, magnetic encoders~\cite{Iqbal2011, Sarakoglou2016, Agarwal2015} or potentiometers~\cite{Sarac2017} can be used for such measurements. Even though the speed and efficacy of finger pose tracking depend on the quality of sensors and capabilities of control board, they can be implemented basically for all kinematics choices and target applications.

\vspace*{-.5\baselineskip}
\section{Discussion} \label{sec:discussion}

In the previous section, we investigated a wide range of hand exoskeletons with respect to their design choices and target applications under $4$ main design aspects: \emph{mobility, mechanical design, actuation} and \emph{operation strategies}. Even though we have already mentioned the best practices of each aspect for generic exoskeletons, we will summarize and highlight them in a more compact form.

\vspace*{-0.3\baselineskip}
\subsection{Mobility}

\noindent \textbf{Hand mobility:} A generic exoskeleton should assist user's natural finger movements. The designer can choose to independently control $5$ fingers or $4$ fingers, while the little finger is either left free or coupled with the ring finger. In particular, the anatomic coupling between the ring and little fingers would allow designers to simplify the mechanical system without sacrificing the natural hand movements.
\vspace*{0.3\baselineskip}

\noindent \textbf{Finger mobility:} A generic exoskeleton should allow finger joints to flex/extend in different synergies, based on different tasks. The designer can passively abduct/adduct MCP joint, since it does not significantly change the task performance during ADLs. Furthermore, the designer might focus on flexion/extension of MCP and PIP joints only, since the natural coupling between DIP and PIP joints would cause the DIP joints to move accordingly even without assistance. The designer can choose to achieve $2~DoF$ or $3~DoF$ mobility for each finger either by controlling them independently, or by coupling them using strategies to adjust for different tasks.
\vspace*{0.3\baselineskip}

\noindent \textbf{Interaction points:} The number of interaction points between a generic exoskeleton and user's fingers should be decided according to the number of finger mobility.


\vspace*{-0.3\baselineskip}
\subsection{Mechanical design}

Kinematics selection should be made based on mobility. The designer can adopt glove-based or linkage-based exoskeletons for independently controlled finger components. Despite their bulky and expensive design, they will achieve high performance for strict trajectory following tasks. Furthermore, mechanical and finger joints must be aligned carefully to ensure user's safety and efficacy of applied forces.

The designer can also couple finger joints using contact based underactuation, such that a single actuator moves multiple finger joints while passive elements adjust the operation based on interaction forces acting on finger phalanges. Thanks to the automatic adjustability, the mechanism can be simplified significantly. Furthermore, passive elements ensure users' safety during operation. However, they require complex control strategies to achieve high tracking performance.

In either case, finger components of generic linkage-based exoskeletons should be placed on the dorsal side.

\vspace*{-0.3\baselineskip}
\subsection{Actuation}

Devices with independent joint control would have remote actuators with cable transmission. Doing so would allow designers be choose any actuator that satisfies the need for output force and backdriveability. However, user's ability to move in the environment would be limited.

Devices with underactuation can have minimized linear actuators inserted on top of the hand with direct-drive transmission. Direct-drive improves the compactness and portability of the device, while limiting the actuator choice. Linear actuators can control these devices to open/close fingers in a multiple finger implementation. However, minimized linear actuators mostly have mechanical gearboxes, affecting implicit backdriveable.


\vspace*{-0.3\baselineskip}
\subsection{Operation strategies}

\noindent \textbf{Control:} A generic hand exoskeleton should be backdriveable with or without control, depending on the actuator selection. Additional control strategies might be used for different target applications.
\vspace*{0.3\baselineskip}

\noindent \textbf{Finger pose:} For devices with independent joint control, the actuator displacements measure the finger pose directly. For underactuated devices, additional sensors and forward kinematics are needed to estimate the finger pose.

\vspace*{-.5\baselineskip}
\section{Conclusion}

Thanks to current technological trends and clinical studies, rehabilitation applications can no longer be considered independently from assistive or haptic applications. For example, an exoskeleton can be used for physical rehabilitation, where therapy exercises require patients to interact with real objects or allow them to complete certain tasks virtually shown during serious game scenarios. With the evolving use for hand exoskeletons, future designers should adopt the most efficient hand exoskeleton designs. Such designs can be possible only after studying the design choices of the current devices and their impacts on the applications.

In this paper, we investigated a wide range of hand exoskeletons existing in the literature based on different design aspects, from the perspective of target applications. This investigation showed that most of the exoskeletons are specifically designed for a single application, and cannot be extended for others. In particular, each application requires hand exoskeletons to satisfy certain requirements and a generic exoskeleton must satisfy all of them. We defined a set of design selections that might lead designers to cover each requirement. We detailed the possible design choices for each selection and highlighted the ones that can be used for a generic exoskeleton. In the end, we also summarized the best practices while designing a new device.


We would like to make a note about the data gathering for this paper. We noted a lack of quantitative data across many of the publications surveyed here. This lack of data prevented us from performing a quantitative analysis, and thus offering strong statements for related design selections. Instead, we were only able to make less precise generalizations based on qualitative inferences (see Section~\ref{sec:assumptions}). 
We believe that future reviewers would be well served if our colleagues started reporting such quantitative measures, as this can only strengthen our knowledge of the existing devices and our production of future innovations.

In spite of these limitations, we are able to conclude that a generic exoskeleton can be designed with either independent finger control~\cite{Kim2017}, or contact-based underactuation~\cite{Sarac2016}. We should also mention that some studies have promising kinematics structures, but they need to increase the number of independent fingers to fit the requirements of a generic exoskeleton~\cite{Hasegawa2011, Burton2011}. In addition, some of the finger exoskeletons are suitable for a generic exoskeleton if implemented in a multi-finger fashion~\cite{Wang2009, Jones2010, Polotto2012, Hocaoglu2009}. Nevertheless, the search for the most efficient device is not over yet. We hope that this literature survey will provide useful guidelines and practices for future designers while creating new, efficient generic hand exoskeletons.

\begin{table*}[htb]
\scriptsize
\caption{\textbf{5 fingered hand exoskeletons}: Main application (rehabilitation (R) / assistive (A) / haptic (H)), number of independent fingers, number of assisted and independent DoF for each finger, mechanism type, device placement (dorsal (DOR)/ palmar (PAL)/ lateral (LAT)), actuator, control modes (position (POS)/ velocity (VEL)/ backdriveable (BAC)/ admittance (ADM)/ impedance (IMP)/ EMG triggered (EMG)), pose estimation method (encoder (ENC)/ flex sensor (FLE)/ motion tracking (MT)/ additional sensor (SEN))}
\vspace*{-2\baselineskip}
\label{tab:5fin}
\begin{center}
{\rowcolors{2}{red!20!red!10}{red!5!red!2}
 \renewcommand\arraystretch{1.2}
 \begin{tabular}{p{3.5cm}| p{.6cm} p{0.9cm} p{1.75cm} p{2.2cm} p{1.4cm} p{1.6cm} p{1.5cm} p{1.2cm} }
 \hline
Device & Main App. & Indep. Fing. & Fing. DoF (active) & Mechanism & Placement & Actuation & Control modes & Fing. pose \\ \hline 
\textbf{Tong et al.}~\cite{Tong2010, Tong2013, Tong2015} & R & 5 & 2(1) & Link (coupled) & DOR & DC (Lin) & BAC, EMG & ENC \\ 
\textbf{HEXOSYS}~\cite{Iqbal2011, Iqbal2015} & R & 5 & 4(1) & Link (fingertip) & DOR & DC (Rot) & POS & SEN \\ 
\textbf{SAFE}~\cite{BenTzvi2015, Ma2016} & H,R & 5 & 3(1) & Link (fingertip) & DOR & DC (Rot) & ADM & SEN \\
\textbf{Fu et al.}~\cite{Fu2007} & R & 5 & 4(1) & Link (coupled) & DOR & DC (Rot) & POS & SEN \\
\textbf{HANDEXOS}~\cite{Chiri2009, Chiri2012} & R & 5 & 4(1) & Link (coupled) & DOR & DC (Rot) & BAC & SEN,MT \\
\textbf{Rahman et al.}~\cite{Rahman2012} & R & 5 & 3(1) & Link (coupled) & DOR & DC (Lin) & POS & ENC \\ 
\textbf{Jo et al.}~\cite{Jo2014, Jo2015} & H & 5 & 3(1) & Link-glove & DOR & DC (Lin) & VEL & FLE \\ 
\textbf{PneuGlove}~\cite{Connelly2010} & H,R & 5 & 3(1) & Glove & DOR, PAL & Pneumatic & POS, BAC & FLE \\ 
\textbf{Fang et al.}~\cite{Fang2009} & H & 5 & 3(1) & Link (fingertip) & DOR & DC (Rot) & ADM & ENC \\ 
\textbf{Delph et al.}~\cite{Delph2013} & R & 5 & 3(1) & Glove & DOR, PAL & Servo & POS, EMG & - \\ 
\textbf{Polygerinos et al.}~\cite{Polygerinos2015} & R & 5 & 3(1) & Link (compliant) & DOR & Pneumatic & POS & - \\ 
\textbf{Cui et al.}~\cite{Cui2015} & R & 5 & 3(1) & Link (coupled) & DOR & DC (Lin) & POS & - \\ 
\textbf{Kim et al.}~\cite{Kim2017} & R & 5 & 3(2) & Link (coupled) & DOR & DC (Rot) & POS & ENC \\
\textbf{Decker et al.}~\cite{Decker2017} & R & 5 & 2(1) & Link-glove & DOR & DC (Rot) & BAC & FLE \\ 
\textbf{Jo et al.}~\cite{Jo2017} & R & 5 & 2(1) & Link-glove & DOR & DC (Lin) & POS & ENC \\ 
\textbf{Yap et al.}~\cite{Yap2015, Yap2017} & A,R & 5 & 3(1) & Compliant & DOR & Pneumatic & POS & - \\ 
\textbf{Lu et al.}~\cite{Lu2016} & R & 5 & 2(1) & Link (coupled) & DOR & DC (Rot) & - & ENC \\ 
\textbf{Sarac et al.}~\cite{Sarac2016, Sarac2017, Sarac2018, Gabardi2018} & R,A,H & 5 & 2(1) & Link (underact.) & DOR & DC (Lin) & POS, ADM & SEN \\ 
\textbf{Hasegawa et al.}~\cite{Hasegawa2008, Hasegawa2011} & A & 3 & 3(3) & Link (indep) & DOR, LAT & DC (Rot) & POS, ADM & ENC \\ 
\textbf{Burton et al.}~\cite{Burton2011} & R & 3 & 3(2) & Link (coupled) & DOR & Pneumatic & - & ENC,MT \\ 
\textbf{Ferguson et al.}~\cite{Ferguson2018} & R & 3 & 4(2) & Link (coupled) & DOR & DC (Rot) & POS, ADM & - \\ 
\textbf{BiomHED}~\cite{Lee2013, Lee2014} & R & 2 & 4(3) & Link-glove & DOR, PAL & DC (Rot) & ADM & MT \\ 
\textbf{BRAVO Hand}~\cite{Troncossi2012, leonardis2015emg} & R & 2 & 3(1) & Link (mitten) & DOR & DC (Rot) & POS, EMG & ENC \\ 
\textbf{Mulas et al.}~\cite{Mulas2005} & R & 2 & 3(1) & Link-glove & DOR, LAT & Servo & EMG & ENC \\ 
\textbf{HANDSOME}~\cite{Brokaw2011} & R & 2 & 1 & Link (indep) & DOR & Spring & - & ENC \\ 
\textbf{Exophalanx-2}~\cite{Kobayashi2013} & H & 2 & 3(2) & Link-glove & DOR, LAT & SMA & ADM & ENC \\ 
\textbf{Li et al.}~\cite{Li2017} & R & 1 & 3(1) & Glove & DOR & Pneumatic & POS & FLE \\ 
\end{tabular} }
\vspace*{-1.5\baselineskip}
\end{center}
\end{table*}

\begin{table*}[htb]
\scriptsize
\caption{\textbf{4, 3, 2 and 1 fingered hand exoskeletons}: Main application (rehabilitation (R) / haptic (H) / assistive (A)), number of assisted and independent fingers, number of assisted and independent DoF for each finger, mechanism type, device placement (dorsal (DOR)/ palmar (PAL)/ lateral (LAT)), actuator, control modes (position (POS)/ velocity (VEL)/ backdriveable (BAC)/ admittance (ADM)/ impedance (IMP)/ EMG triggered (EMG)), pose estimation method (encoder (ENC)/ flex sensor (FLE)/ motion tracking (MT)/ additional sensor (SEN))}
\vspace*{-2\baselineskip}
\label{tab:4321fin}
\begin{center}
 \renewcommand\arraystretch{1.2}
{\rowcolors{2}{red!20!red!10}{red!5!red!2}
 \begin{tabular}{p{3cm}| p{.6cm} p{1.4cm} p{1.75cm} p{2.2cm} p{1.4cm} p{1.6cm} p{1.5cm} p{1.2cm} }
 \hline
Device & Main App. & Assist. Fing. (Indep.) & Fing. DoF (active) & Mechanism & Placement & Actuation & Control modes & Fing. pose \\ \hline 
\textbf{Rutgers Master}~\cite{Bouzit2002} & H & 4(4) & 4(1) & Link (fingertip) & PAL & Pneumatic & POS & SEN \\ 
\textbf{Popov et al.}~\cite{Popov2017} & A & 4(4) & 3(1) & Glove & DOR & Pneumatic & POS & FLE \\ 
\textbf{Allotta et al.}~\cite{Allotta2015} & A & 4(4) & 3(1) & Link (coupled) & DOR & Servo & POS & ENC \\ 
\textbf{Wu et al.}~\cite{Wu2010} & R & 4(1) & 2(2) & Link (mitten) & DOR & Pneumatic & POS & SEN \\ 
\textbf{Weiss et al.}~\cite{Weiss2013} & R & 4(1) & 4(1) & Link (coupled) & DOR,LAT & DC (Rot) & POS & SEN \\
\textbf{Wei et al.}~\cite{Wei2013} & R & 4(1) & 2(1) & Link (mitten) & DOR,LAT & DC (Rot) & - & FLE \\
\textbf{Arata et al.}~\cite{Arata2013, Arata2016} & A & 4(1) & 3(1) & Link (compliant) & DOR & DC (Lin) & POS, BAC & ENC,MT \\ 
\textbf{Gasser et al.}~\cite{Gasser2015, Gasser2017} & R,A & 4(1) & 2(1) & Link (mitten) & DOR,LAT & DC (Rot) & POS & ENC \\ 
\textbf{Lince et al.}~\cite{Lince2017} & R,A & 4(1) & 3(1) & Link (mitten) & DOR & DC (Rot) & EMG & ENC \\ \hline \hline
\textbf{Wei et al.}~\cite{Wei2017} & R,A & 3(3) & 3(2) & Link (coupled) & DOR & DC (Rot) & IMP & ENC \\ 
\textbf{Exophalanx}~\cite{Kobayashi2012} & H & 3(3) & 3(1) & Link-glove & DOR & SMA & ADM & FLE,MT \\
\textbf{Ryu et al.}~\cite{Ryu2008} & H & 3(3) & 3(1) & Link (compliant) & DOR & Pneumatic & POS, BAC & FLE \\ 
\textbf{Sarakoglou et al.}~\cite{Sarakoglou2016} & H & 3(3) & 4(1) & Link (fingertip) & DOR & DC (Rot) & ADM & SEN \\ 
\textbf{In et al.}~\cite{In2010, In2015} & A & 3(1) & 3(1) & Glove & DOR & DC (Rot) & ADM & ENC \\ 
\textbf{Geo et al.}~\cite{Geo2016} & R & 3(1) & 3(1) & Link (coupled) & DOR & DC (Rot) & - & ENC \\ \hline \hline 
\textbf{Cempini et al.}~\cite{Cempini2013, Cortese2015} & R & 2(2) & 4(1) & Link (f.coupled) & DOR,LAT & DC (Rot) & VEL & ENC \\ 
\textbf{FINGER}~\cite{Taheri2014} & R & 2(2) & 2(2) & Link (indep) & DOR & DC (Lin) & POS, BAC & ENC \\ 
\textbf{WHIPFI}~\cite{Gosselin2005} & H & 2(2) & 3(1) & Link (fingertip) & DOR & DC (Rot) & ADM & SEN \\ 
\textbf{Fontana et al.}~\cite{Fontana2009} & H & 2(2) & 4(3) & Link (coupled) & DOR & DC (Rot) & - & ENC \\ 
\textbf{iHandRehab}~\cite{Li2011} & R & 2(2) & 4(4) & Link (indep) & DOR & DC (Rot) & POS, BAC & ENC \\ 
\textbf{Fiorilla et al.}~\cite{Fiorilla2008} & R & 2(2) & 1 & Link (MCP) & DOR & DC (Rot) & - & SEN \\ \hline \hline
\textbf{Wang et al.}~\cite{Wang2009} & R & 1(1) & 4(4) & Link (indep) & DOR & DC (Rot) & POS, BAC & ENC \\
\textbf{Yamaura et al.}~\cite{Yamaura2009} & R & 1(1) & 3(2) & Link (coupled) & DOR & Servo & POS & ENC \\
\textbf{Jones et al.}~\cite{Jones2010, Jones2012} & R & 1(1) & 3(3) & Link (indep) & DOR,LAT & DC (Rot) & POS, BAC & ENC \\ 
\textbf{Polotto et al.}~\cite{Polotto2012} & R, A & 1(1) & 4(4) & Link (indep) & DOR,LAT & DC (Rot) & POS, BAC & ENC \\ 
\textbf{Tang et al.}~\cite{Tang2011, Tang2013} & R & 1(1) & 2(1) & Link (coupled) & DOR & Ultrasonic & POS, BAC & ENC \\ 
\textbf{DiCicco et al.}~\cite{DiCicco2004} & A & 1(1) & 3(1) & Link (coupled) & DOR & Pneumatic & EMG & ENC \\ 
\textbf{Wege et al.}~\cite{Wege2005, Wege2006, Wege2007} & R & 1(1) & 4(1) & Link (coupled) & DOR & DC (Rot) & POS, EMG & SEN \\ 
\textbf{Sun et al.}~\cite{Sun2009} & R & 1(1) & 4(1) & Link (fingertip) & DOR & Pneumatic & POS, ADM & MT \\
\textbf{Agarwal et al.}~\cite{Agarwal2013} & R & 1(1) & 4(3) & Link (indep) & DOR & Pneumatic & - & SEN \\ 
\textbf{Agarwal2 et al.}~\cite{Agarwal2015} & R & 1(1) & 3(2) & Link (coupled) & DOR & Elastic & IMP & ENC\\ 
\textbf{AssistOnFinger}~\cite{Hocaoglu2009, Ertas2014} & R & 1(1) & 3(1) & Link (underac.) & DOR & DC (Rot) & POS,BAC,EMG & - \\
\textbf{Aubin et al.}~\cite{Aubin2013} & R & 1(1) & 2(2) & Link-glove & DOR & Servo & POS, BAC & ENC \\ 
\textbf{Lee et al.}~\cite{Lee2015} & R & 1(1) & 3(1) & Link (coupled) & DOR & DC (Rot)& - & SEN \\ 
\textbf{Maeder York et al.}~\cite{Maeder-York2014} & R & 1(1) & 3(1) & Link (compliant) & DOR & Pneumatic & - & ENC \\ 
\end{tabular} }
\vspace*{-1.5\baselineskip}
\end{center}
\end{table*}

\vspace{-3mm}

\small
\bibliographystyle{IEEEtran}
 \bibliography{HandExoskeletonsLiterature_Final}

\begin{IEEEbiography}[{\includegraphics[width=1in,height=1.25in,clip,keepaspectratio]{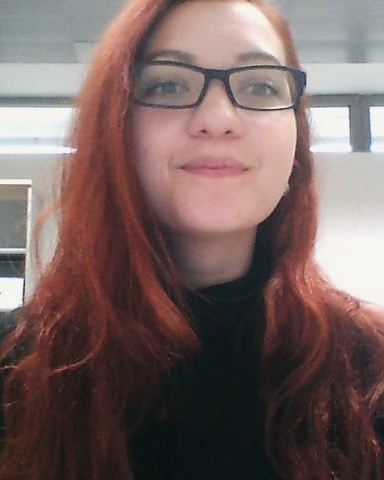}}]{Mine Sarac}
Mine Sarac graduated with a B.Sc. degree in Electrical and Electronics Engineering from Yeditepe University, Istanbul, Turkey in 2011 and M.Sc. degree in Mechatronics from Sabanci University, Istanbul, Turkey in 2013. She received her PhD in Perceptual Robotics from Scuola Superiore Sant'Anna, Pisa, Italy in 2017. Currently, she is working as a post-doc student as a part of Charm Lab in Mechanical Engineering at Stanford University. Her research interests are rehabilitation robotics, human-machine interaction, haptic systems and compliant mechanisms. She is a member of the IEEE.
\end{IEEEbiography}

\begin{IEEEbiography}[{\includegraphics[width=1in,height=1.25in,clip,keepaspectratio]{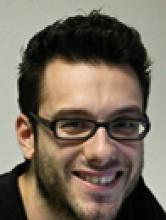}}]{Massimiliano Solazzi}
Massimiliano Solazzi received the PhD degree in innovative technologies from Scuola Superiore Sant’Anna, in 2010. He is an assistant professor in applied mechanics with the Scuola Superiore Sant’Anna, Pisa, Italy. He carries out his research at the PERCRO Laboratory-TeCIP. His research interests concerns: the design of robotic interfaces for virtual reality, teleoperation and rehabilitations, and the psychophysical validation of HMI. He is a member of the IEEE.
\end{IEEEbiography}


\begin{IEEEbiography}[{\includegraphics[width=1in,height=1.25in,clip,keepaspectratio]{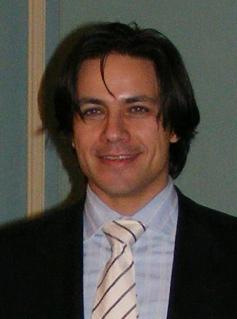}}]{Antonio Frisoli}
Antonio Frisoli received the MSc degree in mechanical engineering, in 1998, and the PhD degree (with Hons.) in industrial and information
engineering from Scuola Superiore Sant’Anna, Pisa, Italy in 2002. He is a full professor of mechanical engineering with Scuola Superiore Sant’Anna, where he is currently head of the HRI area at PERCRO Laboratory-TeCIP and former chair of the IEEE Technical Committee on Haptics. His research interests concern the design and control of haptic devices and robotic systems, rehabilitation robotics, advanced HRI, and kinematics. He is a member of the IEEE.
\end{IEEEbiography}

\end{document}